\documentclass[letterpaper]{article} % DO NOT CHANGE THIS
\usepackage{aaai25}  % DO NOT CHANGE THIS
\usepackage{times}  % DO NOT CHANGE THIS
\usepackage{helvet}  % DO NOT CHANGE THIS
\usepackage{courier}  % DO NOT CHANGE THIS
\usepackage[hyphens]{url}  % DO NOT CHANGE THIS
\usepackage{graphicx} % DO NOT CHANGE THIS
\usepackage{natbib}  % DO NOT CHANGE THIS AND DO NOT ADD ANY OPTIONS TO IT
\usepackage{caption} % DO NOT CHANGE THIS AND DO NOT ADD ANY OPTIONS TO IT
\usepackage{algorithm}
\usepackage{algorithmic}
\usepackage{booktabs} 
\usepackage{makecell}
\usepackage{svg}
\usepackage{amsmath,amsfonts}
\usepackage{subcaption}

\def\gE{{\mathcal{E}}}

\def\gG{{\mathcal{G}}}

\def\gN{{\mathcal{N}}}

\def\gR{{\mathcal{R}}}

\def\gV{{\mathcal{V}}}

\usepackage{newfloat}
\usepackage{listings}
\DeclareCaptionStyle{ruled}{labelfont=normalfont,labelsep=colon,strut=off} % DO NOT CHANGE THIS
\floatstyle{ruled}
\newfloat{listing}{tb}{lst}{}
\floatname{listing}{Listing}
\title{Investigating Relational State Abstraction in Collaborative MARL}
\author{
    Sharlin Utke\textsuperscript{\rm 1}, 
    Jeremie Houssineau\textsuperscript{\rm 2}, 
    Giovanni Montana\textsuperscript{\rm 1}
}
\affiliations{
    \textsuperscript{\rm 1}University of Warwick, Coventry CV4 7AL, United Kingdom \\
    \textsuperscript{\rm 2}Nanyang Technological University, 50 Nanyang Avenue, Singapore 639798 \\
    sharlin.utke@warwick.ac.uk, jeremie.houssineau@ntu.edu.sg, g.montana@warwick.ac.uk
}

\usepackage{bibentry}
\begin{document}

\maketitle

\begin{abstract}
This paper explores the impact of relational state abstraction on sample efficiency and performance in collaborative Multi-Agent Reinforcement Learning. The proposed abstraction is based on spatial relationships in environments where direct communication between agents is not allowed, leveraging the ubiquity of spatial reasoning in real-world multi-agent scenarios. We introduce MARC (Multi-Agent Relational Critic), a simple yet effective critic architecture incorporating spatial relational inductive biases by transforming the state into a spatial graph and processing it through a relational graph neural network. The performance of MARC is evaluated across six collaborative tasks, including a novel environment with heterogeneous agents. We conduct a comprehensive empirical analysis, comparing MARC against state-of-the-art MARL baselines, demonstrating improvements in both sample efficiency and asymptotic performance, as well as its potential for generalization. Our findings suggest that a minimal integration of spatial relational inductive biases as abstraction can yield substantial benefits without requiring complex designs or task-specific engineering. This work provides insights into the potential of relational state abstraction to address sample efficiency, a key challenge in MARL, offering a promising direction for developing more efficient algorithms in spatially complex environments.
\end{abstract}

\begin{links}
\link{Code \& extended version with appendix}{https://github.com/sharlinu/MARC}
\end{links}

\section{Introduction}

Multi-Agent Reinforcement Learning (MARL) has emerged as an extension of single-agent RL, where multiple agents simultaneously interact with the environment to derive their optimal behavior by trial and error. Despite the complexity and dynamics of the learning environment, MARL holds significant promise for modeling real-world systems involving nuanced interactions between multiple entities, such as autonomous vehicle coordination \cite{AutoVehicles2016}, traffic flow optimization \cite{Airtraffic2010}, and team robotics \cite{Multi-robot2012}. These applications often involve collaborative or competitive dynamics that single-agent RL struggles to capture adequately. However, the increase in dimensionality over state and action spaces, the additional agent interactions, and the information influx this entails make sample efficiency a key challenge. 

A critical aspect to sample efficiency is how agents represent the information of what they observe. The observation received by an agent can hold a lot of information, but not all of it is necessary to make an optimal decision. The ability to find abstract features allows for reasoning on a higher, conceptual level and adds robustness to small, task-irrelevant changes; a common principle for good representations \cite{Bengio2012}. Abstraction leverages the underlying structure of the problem to focus on relevant information while reducing its complexity. That way, agents can learn optimal policies with fewer interactions, leading to improved sample efficiency and knowledge transfer to new situations \cite{StructureRL}. 

A natural structure to present the environment is the decomposition into objects and their relations. Recent work in both deep single-agent RL and MARL has demonstrated the benefits of leveraging this relational structure as a graph-based representation in the learning architecture \cite{Bapst2019, GTG2021, informal, Argarwal2020}, improving sample efficiency and generalization capabilities. This seems especially important in MARL, where the complexity often scales with the number of agents. The ability of Graph Convolutional Networks (GCNs) \cite{Scarselli2009,Welling2016,gilmer2017} to model systems where the relationships between entities are critical has brought them to the forefront in many fields of multi-agent systems, ranging from modeling behavior and trajectories in multi-agent systems \cite[e.g.][]{NRI2018, RFM2018, Evolvegraph2020, Kipf2020} to enhancing communication \cite[e.g][]{MAGIC,DGN2020,sriac2021}. 

In this study, we investigate the integration of a simple yet effective relational abstraction to MARL using the architectural flexibility of GCNs. We focus on collaborative tasks with high spatial complexity where direct communication between agents is not permitted due to cost or security constraints, e.g., in underwater robotics \cite{underwater}. We specifically emphasize spatial relationships, as these are ubiquitous and readily available in real-world multi-agent scenarios. Spatial relations provide fundamental information about the relative positions and orientations of agents and objects, which is crucial for navigation, coordination, and collaboration in physical environments. 
This focus allows us to explore how relational information can be leveraged through implicit information already present in the environment, without relying on explicit communication channels or complex architectural designs. Our research aims to address two key questions: (1) Can the incorporation of a state abstraction using spatial inductive biases improve sample efficiency and asymptotic performance in MARL? (2) How do different choices in the design impact the learning in such relational architectures?

To address these questions, we propose MARC (Multi-Agent Relational Critic), a simple multi-agent actor-critic architecture that abstracts the observation based on the relative positions of the entities into a graph-based representation. MARC utilizes a shared relational component within the critic architecture to efficiently learn a structured representation, further aiding sample efficiency. 
We conduct comprehensive empirical evaluation against state-of-the-art (SOTA) MARL baselines and across different tasks including a newly created collaborative multi-agent environment, designed to be a spatially demanding task between heterogeneous agents. Ultimately,  we examine the impact of different design choices in our relational component in Section~\ref{sec:ablation}.

\section{Related work}

\subsubsection{State Abstraction in RL}

Abstraction has been widely studied, with early work showing theoretical properties in single-agent RL \cite{li2006towards, abel2022theoryabstractionreinforcementlearning}. Inspired by \citet{Zucker} and focusing on state abstraction, we define abstraction as a mapping of the ground-truth representation of a state to a simpler, more compact representation by preserving desirable properties and removing less critical information. In other words, abstraction simplifies the information representation by dropping the information that is not essential to the task. Many methods have shown the success of embedding an abstract representation. For example, \citet{Kipf2020} factorize state inputs into objects and apply a relational, object-centric state abstraction to model a multi-object system. \citet{Zhang2021} aim to learn an abstract state representation from high-dimensional observations based on the behavioral similarity between states to encode only information relevant to the task. \citet{Abdel2024} reduce the computational complexity between communicating agents by learning a state abstraction based on quadtree decomposition \cite{quadtree}. \citet{sriac2021} use a state abstraction component in the MARL setting to reduce the high-dimensional observations into a more compact latent presentation using dense neural networks. While the abstraction method and assumption we take are different, we leverage these methods' underlying idea to discard any information irrelevant to the task to create a more compact and efficient representation.

% - Nevertheless, the performance of our algorithms relies on the representation of the data and its features (Bengio).
%- Alternatively,  we can induce the prior knowledge directly into the network architecture.

\subsubsection{Relational Representation in RL}
%, where the type of structure can vary. In RL, it can be considered some additional information on how we can decompose elements of the decision process such as the state, the actions or the value function. The type of structure can vary from possible factorization, a richer latent representation or relational structure if we have information about how the factors in the space relate to each other. For example, as mentioned above, QMIX and VDN are defined on the structural assumption that one can factorize the value functions of the agents, hence aiming to reduce the curse of dimensionality. As MAS exerts a relational compositional structure almost by nature alone we want to focus on enhancing prior knowledge of the relational structure in the RL pipeline. 
%Abstraction patterns are defined on structural information that can be leveraged to reduce the complexity of a problem. 
Before the integration of deep learning, traditional RL methods often falter in environments with relational structures or when generalization beyond initial training conditions is necessary. Relational RL addresses these challenges by learning the optimal policies over the objects and relations using a relational representation such as first-order logic. Whilst this approach has shown improved generalization and scalability, both in single-agent RL \cite{RRL2001, sanner2009, driessens2001} and in MARL \cite{RMARL2006, CommRMARL2010, DILP2022}, the use of first-order logic comes with constraints, such as the need to hand-engineer features \cite{Garnelo2016}. Contemporary methods tackle this issue by learning the relations between objects using deep learning methods \cite{Garnelo2016, GTG2021, zambaldi2018}. These methods assume that observations comprise entities and the relationships between them while using deep learning methods as an inductive bias to learn over these structures. For example, \citet{zambaldi2018} learns the importance of non-spatial relations between entities using attention mechanisms \cite{Vaswani2017}. They show superior performance and generalization capabilities compared to purely local relations in single-agent RL. \citet{GTG2021} connect entities of a grid with a broader set of spatial relations, including remote relations, into a heterogeneous graph and passes them through a Relational Graph Convolutional Network (R-GCN) \cite{Schlichtkrull2018}. Their findings indicate that the imposition of structure by inducing a spatial bias can lead to improved asymptotic performance and generalization capabilities in single-agent tasks. However, they treat every cell in the grid as an entity, whether or not it contains an environment object. We extend this idea to MARL by employing a relational representation between agents and objects in the environment where the importance of the induced relations is implicitly learned using R-GCNs. Additionally, we only consider the environment objects, i.e. agents and other objects, as entities and use fewer relations, proposing a lean and abstract representation that better aligns with the computational complexity of MARL and works on continuous domains as well.
%This means that for a grid of size $n \times n$, they create $n^2$ entities, even if many of these grid cells are empty or irrelevant to the task at hand.
\subsubsection{Relational Inductive Bias in MARL}

Relational inductive bias, loosely defined as the imposition of structural constraints on the learning process based on the relationship between objects \cite{inductivebias}, is a principle commonly embedded in MARL architectures.

A natural way to leverage structure in MARL is on the agent level. In fact, many of the commonly known MARL algorithms own an architecture that represents a form of relational inductive bias based on the structure between agents. For example, value decomposition methods such as QMIX \cite{QMIX} assume conditional independence between the agents to decompose their value function. MAAC \cite{MAAC} assumes that the influence of one agent's information on another can vary. Each critic can dynamically select which agents to focus on by assessing the relevance of the encoded information from other agents via a multi-head attention layer that is shared between critics. The actor-critic methods introduced by \citet{g2anet} extend the constraint imposed in MAAC by using an additional hard-attention layer to strengthen 
the assumption that not all other agents' information is relevant to succeed, further reducing the complexity of the game dynamics. \citet{GPG2019} leverage the underlying graph structure and symmetry between large numbers of homogeneous agents to parameterize the policies using a GCN framework. 

Whilst the decomposition of the MARL architecture on the agent level is very intuitive, some methods leverage the structure already found within the observation. For instance, VMARL \cite{VMARL} transforms high-dimensional visual inputs to an object-centric intermediate state representation where environment objects are linked by their proximity, before being fed to policy and actor networks. MAGNet \cite{MAGNET19} considers all environment objects as entities when pre-training a static relevance graph with known node and edge types, which is then used to represent the observations. \citet{Argarwal2020} and \citet{informal} employ distance-based observation graphs with learned attention weights between agents and objects, demonstrating that graph representations of the environment allow for a framework invariant to permutation and the number of entities in the environment. They both assume shared rewards among homogeneous agents that allow for communication within their neighborhood. What most of these methods have in common is that they take advantage of the strong relational inductive bias posed by graph neural network architectures, enforcing learning over entities and their relations. We similarly leverage graph neural architectures to enforce a structured observation. However, we leverage inherent spatial symmetries to reduce the observation complexity and employ a MARL architecture also applicable to heterogeneous agents.

%In contrast, we abstract away information about absolute position and distances, reducing the complexity of the representation and making it invariant not only towards permutation but also to translations of the observation. 
% This yields improved sample efficiency and generalization capabilities.

\section{Methodology}
\label{sec:method}

\subsection{Preliminaries}

We work under the framework of partially observable Markov Games, with $S$ being the state space in which each of the $N$ agents has their own action space $A_i$, with  $i = 1, \ldots, N$, forming a joint action space $A = A_1 \times A_2 \times \dots \times A_N$. After taking an action, each agent $i$ receives an observation $o_i \in O_i \subset S$. Moreover, we assume individual reward functions, $R_i: S \times A \to \mathbb{R} $, which gives a reward signal after every step. At each time step, the agents simultaneously choose actions according to their respective policies, $\pi_i: O_i \mapsto P (A_i)$, which depend on the observation they receive. Consequently, the environment changes in line with the transition dynamics $T: S \times A \times S \to [0,1]$ to a new state. The goal is that every agent finds the optimal policy that maximizes their expected cumulative return $J_i (\pi_i) = \sum^T_{t=0} \gamma^t r_i^t$, where $\gamma \in (0,1]$ is the discount factor incorporating uncertainty about future returns, and where $r_i^t$ is the individual reward received at time step $t$.

\begin{figure*}
    \centering
    \includegraphics[width=0.95\textwidth]{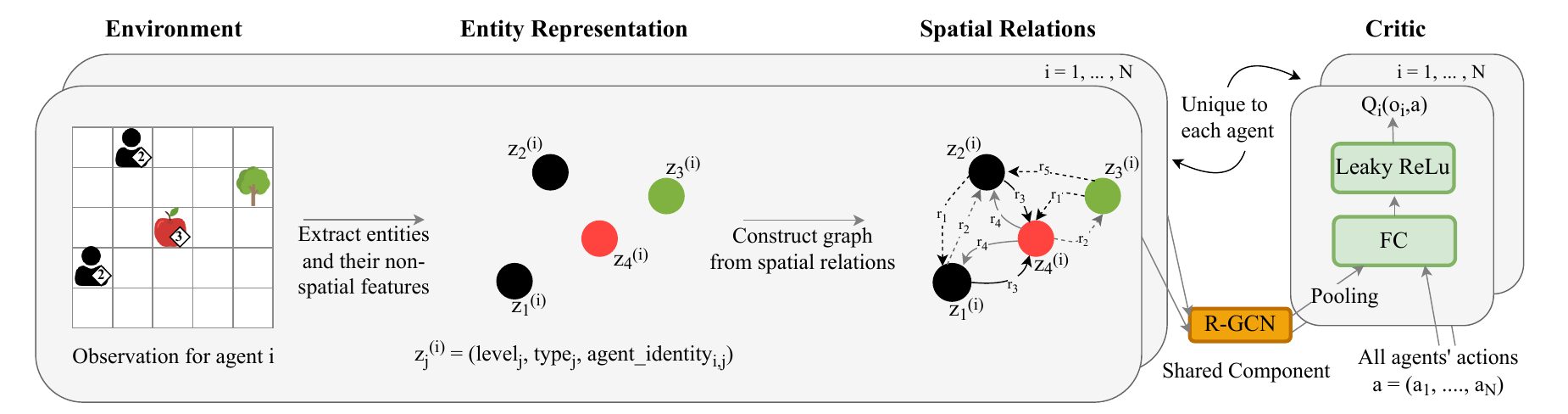}
    \caption{Overview of our MARC architecture on the example of level-based foraging. Without adding information, the observation is constructed as a graph, with objects and agents as entities and our chosen set of relations. We then pass this relational graph into a shared R-GCN component, followed by an individual head for each agent to estimate the state-action value.}
    \label{fig: MARL_RGCN_architecture}
\end{figure*}

\subsection{Abstract Observation Representation}
Our objective is to design a sample-efficient multi-agent actor-critic architecture that decomposes the observations based on spatial inductive biases. We achieve this by employing a form of state abstraction: we simplify the observation representation by dropping information that is not essential to the task. This can also be described as domain reduction where we collapse observations into equivalent clusters, causing some observations to be indistinguishable and ultimately reducing observation complexity \cite{Zucker}.
%A well-known example of such transformation is to coarsen the resolution of an image, where adjacent pixels are averaged together to yield a coarser grid. Similarly, 
%. While an abstraction may not perfectly characterize the entire state space, it aims to represent the most relevant information. 
%To reflect this in our methodology, we collapse the observation space into a simpler, more coarse representation based on relative spatial information. This abstraction reduces the size and complexity of the observation space and clusters similar observations. We propose constructing a spatial graph from the observation and employing a form of GNN to learn over this structured observation.

The abstraction assumption we make is that the relative positioning of entities is relevant, not their absolute positions. We are inducing an equivalence between observations, where we group observations with a similar spatial structure. This induces a translation invariance that applies to both remote and local relations in the observation space. For this to hold, we assume that the relative spatial relations can be extracted from the observation. This type of spatial information is inherent in many common environments and real-world scenarios and offers an intuitive example of using existing structures in the observations.
%Whilst the entities do not necessarily need to be known, we find that assuming their knowledge increases sample efficiency.
% they say semantic knowledge on node features needs to be known and the 2D grid space 

We hypothesize that this abstraction is particularly fruitful in discrete domains. Discrete states can be clearly separated from each other, which makes it easy to exactly define boundaries for any spatial relations. In contrast, continuous state spaces have a higher state complexity, as they are infinite expressions of the state and small changes can have a significant impact on the optimal action. Hence, clustering continuous state spaces can lead to a stronger loss of information in the representation that could impact performance \cite{li2006towards}. Whilst these challenges may affect the effectiveness of our abstraction, we test the robustness of our approach on continuous domains as well. 

Effectively leveraging graph structures to impose such inductive bias in MARL poses the key challenges of (a) determining how entities are represented; (b) finding an informative yet efficient use of relations; (c)
aggregating information across the graph to propagate relevant signals; (d) finding a computationally efficient way of incorporating the structural information into the MARL architecture. In the following, we address these challenges through specific design choices to create a structured, more compact observation representation that leverages the permutation invariance to the order of entities and the translation invariance to the absolute position between entities. An overview of the steps and our overall architecture can be seen in Figure~\ref{fig: MARL_RGCN_architecture}.

\subsubsection{Entity Representation} \label{methods: entities}

In many of the discussed MARL methods, the focus lies on the interaction between agents, where the information they share is usually an encoding of their individual observation and action \cite[e.g.][]{MAAC, g2anet,DGN2020, sriac2021}. We want to emphasize the structure already present within the observation itself. Hence, we aim to find a structured representation of the observation that does not only consider the agents but also all the other objects in the environment.

%In contrast, we construct an entity set V from all agents and objects. 

Typically, observations are given as fixed-sized vectors that contain the positions and attributes of agents and objects. This enforces an artificial ordering between the entities that is not desirable. The structure of such an observation is commonly a design choice and can be varied without great loss of generality. Consequently, we assume that the positions and attributes of the agents and environment objects can be extracted. In detail, we first construct an entity set $\gV$ from all agents and objects. We then take the non-spatial information from all agents and objects, such as their level, type, if they are carrying objects, an identifier or other attributes. This results in a corresponding entity feature matrix $Z \in \mathbb{R}^{d \times |\gV|}$ with $d$ being the number of entity features. 

\subsubsection{Spatial Relations}
Our abstraction assumption is that only the relative spatial information is essential to solve the task. In line with that hypothesis, we do not consider the absolute position $(x, y)$ of an entity as an entity feature but transform this information into relative spatial edges between all entities. These edges are established using spatial predicates $r(a,b) \leftarrow \text{condition}$, such as $\textit{left}(a,b) \leftarrow x_a < x_b$, which indicates that entity $a$ is to the left of entity $b$. Our chosen set of relations $\gR = \{  {\textit{left}, \textit{right}, \textit{top}, \textit{bottom}, \textit{adjacent}, \textit{aligned}} \} $ are directed edges designed to balance sufficient expressiveness with computational efficiency. An evaluation of this selection along with the potential impact of additional relations, are discussed in Section~\ref{sec:ablation}. Since some entities, such as the agents, are dynamic, the spatial relations between entities change at every time step. Unlike in other methods \cite{informal,Argarwal2020}, the edges do not feature the distance between entities. This allows us to induce translation invariance, building a compact abstraction that treats observations with the same relative spatial structure as equivalent. We refer to the appendix for more details on the invariance of our abstraction.

%We construct our relational graph using a deterministic function $g_{\mathcal{R}}$, which maps agent observations and our relational set $\gR$ to generate relational graphs. For any given observation $o_i$ from agent $i$, the corresponding relational graph is $\gG_i = g_{\mathcal{R}}(o_i)$. This method captures rich spatial information in a structured way, providing a strong foundation for learning and decision-making in multi-agent environments by explicitly modeling spatial relationships. 

\subsubsection{Observation Encoding}

Having established the structure of the relational graph, our next objective is to obtain a higher-level representation of the observation, informed by the spatial relation between the entities. For this, we employ R-GCN \cite{Schlichtkrull2018} updates, chosen for the ability to handle multiple relationship types. It updates entity representations by evaluating the entities' individual features and aggregating information from connecting entities depending on their relation type. 

Formally, our graph is denoted as $\gG = (\gV, \gE, \gR, Z)$, where $\gV$ is the set of all entities, $\gE$ represents directed edges signifying relationships, $\gR$ categorizes the types of these relationships and $Z$ denotes the entity-feature matrix. 
The feature update for each entity $v \in \gV$, initially represented by $z_v \in \mathbb{R}^d$, is governed by $z_{v}' = \sigma\left(\sum_{r \in \gR}\sum_{u \in \gN_{r}(v)} \frac{1}{|\gN_{r}(v)|} W_{r} z_{u} + W_{0} z_{v}\right),$ where $\sigma$ is a non-linear activation function. Each relation type $r$ has an associated weight matrix $W_{r} \in \mathbb{R}^{d' \times d}$, customizing the update to the specific nature of the relationship. An auxiliary weight matrix $W_0 \in \mathbb{R}^{d' \times d}$ integrates the entity's original features. The term $\gN_{r}(v)$ represents the neighboring entities of entity $v$ for a given relation type $r$ and the aggregation is normalized by $|\gN_{r}(v)|$.
%This update mechanism employs an aggregation function that applies relation-specific linear transformations to the features of neighboring entities, according to the type of their connections. 
%To ensure stability and account for variations in the number of neighboring entities, 
After applying a number of R-GCN layers as specified, we obtain an updated feature matrix $Z' \in \mathbb{R}^{d' \times |\gV|}$. To generate a fixed-size representation, we apply a feature-wise max-pooling operation, resulting in observation encodings $e(o_i) = \operatorname{max-pool}(Z')$, where $Z'$ implicitly depends on the original observation $o_i$.

This entire process - from initial graph construction through to the final pooling operation - acts as a unified observation encoder that aligns with common MARL environments. It transforms individual agent observations into a compact, relational representation, emphasizing the understanding of entity relationships while excluding nonessential details. Reducing the details in the representation enhances computational efficiency without sacrificing essential information for decision-making. 

By transforming the observation into a graph representation and employing R-GCN updates followed by max-pooling, we obtain observation encodings that are invariant to the order of the input elements. This is a very desirable property, as there is no natural ordering between the objects in an environment. Furthermore, by fully removing the absolute information on positions and distance between entities, we not only reduce the observation complexity but also induce a translation-invariant representation. 
%Translation shifts of the absolute positions of the environment objects do not influence the relative positioning of the environment objects, leaving the final structured representation unchanged.

%This structured representation is subsequently used by other components of our MARL architecture, enhancing the agents' ability to make informed decisions based on their spatial context within a multi-agent scenario.

%The geometric deep learning blueprint proposed by Bronstein et al. provides a unified framework for understanding various deep learning architectures in terms of symmetries and invariances. Equally, we can use this framework to demonstrate the symmetries present in our system, namely permutation and translation invariance.

%The domain is defined as $\Omega = (\gG, \gE)$. To show permutation invariance to the order of inputs, the symmetry group is defined as $\Sigma_n$, representing all $n!$ possible permutations of the input. This symmetry group can be represented by a permutation matrix P. Following \cite{Bronstein}, a graph-wise function $f$ is permutation invariant if:
%\begin{equation} f(X,A) = f(PX,PAP^T) \end{equation}

% For the case of translation invariance, the domain remains unchanged where the group now consists of translations $T(dx,dy)$. 

% We want to show that
% \begin{equation}
%     F(TX) = T(F(X))
% \end{equation}

% where $T$ is defined as the translational operator $T_\delta(x(v)) = x(v + \delta)$.
 
\subsubsection{Learning Algorithm}

Having encoded the observations into a relational graph, our next step involves feeding the relational representation into the MARL framework. In principle, the observation encoder is agnostic to the backbone MARL algorithm and we include a supporting experiment along with a discussion of this aspect in the appendix. Known for effectively balancing SOTA performance with scalability, we employ the popular centralized training with a decentralized execution regime. This framework enables agents to share information during training while maintaining individual decision-making during execution. To enhance scalability and efficiency even further, we depart from the common practice of feeding observation information from all agents into the critic architecture \cite[e.g.][]{MAAC,informal}. Instead, the individual critic only receives the observation information from its own agent and exchanges information implicitly by collectively learning the parameters of the observation encoder, significantly reducing the input dimensionality to the critic. This shared module is complemented by individual heads for each agent, facilitating efficient learning while preserving the capacity for learning individualized behavior. 

In detail, each agent, indexed by \(i\), maintains its own critic and policy, allowing for distinct reward structures and action spaces. Formally, the critic for each agent is defined as $
Q_{\psi_i}(o_i, a) = f_i(e(o_i), a)$, where \(f_i\) is a dense neural network that processes the encoded observation \(e(o_i)\) together with the collective actions \(a = (a_1, \ldots, a_N)\). Here, \(\psi_i\) includes the parameters from both the shared observation encoder \(e\) and the agent-specific dense layers \(f_i\). The critics are jointly optimized to minimize the following regression loss:
$$
\mathcal{L}_Q(\psi) = \sum_{i=1}^N \mathbb{E}_{(o_i, a, r_i, o'_i) \sim D} \left[ (Q_{\psi_i}(o_i, a) - y_i)^2 \right],
$$
where the target is defined as $ y_i = r_i + \gamma \mathbb{E}_{a' \sim \pi_{\bar{\theta}}} \left[ Q_{\bar{\psi}_i}(o'_i, a') - \alpha \log \pi_{\bar{\theta}_i}(a'_i | o'_i) \right]$.
Here, \(\gamma\) is the discount factor and \(D\) represents the replay buffer. Following the soft actor-critic updates \cite{SAC}, we define \(\bar{\psi}_i\) and \(\bar{\theta}_i\) as the target critic and policy parameters for each agent, respectively, and \(\alpha\) as the temperature parameter balancing entropy and reward maximization. The joint target policy vector \(\pi_{\bar{\theta}} = (\pi_{\bar{\theta}_1}, \ldots, \pi_{\bar{\theta}_N})\) comprises policies, each a dense neural network. The individual policies are learned via gradient ascent as in the SAC \cite{SAC} framework and as described in \cite{MAAC} without major modification. The implementation details and hyperparameters can be found in the appendix.

\section{Experiments}

In this section, we detail the experimental setup designed to evaluate the performance and capabilities of our proposed algorithm. We first describe the environments chosen, which are tailored to challenge and showcase the algorithm's spatial reasoning and collaborative capabilities in relationally complex tasks. We then outline the baseline algorithms against which we compare our approach to understand the added value of the relational inductive bias, followed by a comparative analysis of our results.

\subsection{Environments}

We hypothesize that the introduced abstraction learns effectively in spatially complex coordination tasks with sparse rewards. On this basis, we designed a new highly collaborative environment that requires coordination between different types of agents and several object types. Furthermore, we select other collaborative grid environments as they naturally challenge the algorithm's capabilities in spatial reasoning and cooperation under sparse rewards.  

Whilst the employed state abstraction is particularly well suited to the discrete nature of these environments, we also evaluate our method on a continuous domain. Following is an overview of the chosen environment and for further details, we refer to the appendix.

\textbf{Collaborative Pick and Place} (CPP) is a new, collaborative environment with two types of agents that need to pick up and drop off a box at a designated goal, entailing heterogeneous, collaborative agents. Only the picker agents can collect a box whereas the delivery agents can only receive a box and drop it at at a goal location. Once a box is dropped at the goal location, no other box can be placed there. At the beginning of each episode, the boxes, agents and goals are randomly spawned on the grid. Depending on their role, agents receive a reward for successful pick-ups, passes, and drop-offs, as well as for prompt completion of the task. In our experiments, we test the challenging setting of a $10 \times 10$ grid, 2 picker agents, 2 delivery agents and 3 objects\footnote{We made the environment code available at https://github.com/gmontana/CollaborativePickAndPlaceEnv}.

\textbf{Level-based Foraging} (LBF) \cite{lbf2020} situates agents in a grid world where they are rewarded for collecting fruits. As opposed to the original LBF environment, we assume that fruits are on trees that remain on the grid after the fruit has been collected, with a value of $-1$. This alteration demands a higher relational reasoning capability from agents, as they must now navigate around the trees, recognizing them as noncollectable obstacles. For testing high cooperation, our experiments run on a $10\times10$ grid with 4 agents and 4 foods, enforcing cooperation (denoted as 10x10-4a-4f-coop).  To assess scalability, we extend the environment to a $15 \times 15$ grid with 8 agents and 1 fruit (denoted as 15x15-8a-1f-coop).

In \textbf{Wolfpack} \cite{GPL2023}, 3 agents are placed in a $10\times10$ grid to capture 2 prey. In a departure from the original setup, we have introduced sparse rewards by removing additional rewards based on the proximity to prey, significantly weakening the learning signal.

The \textbf{Target} task, based on the multi-agent particle environment \cite{MADDPG} and modified by \citet{informal}, is a continuous domain environment where a number of agents try to reach their target landmarks while avoiding collision with obstacles and other agents.

\subsection{Baselines}

In our study, we choose the baselines based on the following criteria: performance, reproducibility, ability to handle discrete action spaces and similarity to our approach. Following is an overview of the selected baselines; implementation details and hyper-parameter selection can be found in the appendix.

\paragraph{MAAC} \cite{MAAC} also uses SAC as the base RL algorithm. The use of attention between agents represents a different form of relational inductive bias on agent interaction rather than our object-centric representation.
%\item \textbf
\paragraph{GA-AC} is the AC algorithm that makes use of the G2ANet mechanism \cite{g2anet}. It builds on MAAC with an additional hard attention layer, which allows for an even more nuanced differentiation of information from other agents and represents an even stronger inductive bias than MAAC. 
%It also targets the same goal of simplifying the game representation by abstracting it into a simplified version.

%\item \textbf
\paragraph{InforMARL} \cite{informal} introduces a distance-based graph representation of objects and agents that informs policy and critic networks, yielding a similarly structured observation encoding. Tailored to the multi-agent particle environment, it provides a relevant baseline for our experiments in the continuous domain.

%\item \textbf
\paragraph{QMIX} \cite{QMIX} leverages the structural assumption of conditional independence between agents' value functions to factorize it, yielding a rigorously implemented and strong baseline for comparison.

%\item \textbf
\paragraph{MAA2C} \cite{papoudakis2020benchmarking} is an on-policy approach that learns a centralized critic from joint observations without other agents’ actions. It serves as a fast and strong baseline due to its absence of relational inductive bias, meaning it does not explicitly consider relationships between agents or entities in the environment.

\paragraph{MAPPO} \cite{MAPPO} is an extension of single-agent PPO \cite{PPO}, noted for its performance and, similar to MAA2C, does not incorporate a relational inductive bias. It enhances sample efficiency through multiple updates on batches of training data.

\begin{figure*}
\centering
\includegraphics[width=\textwidth]{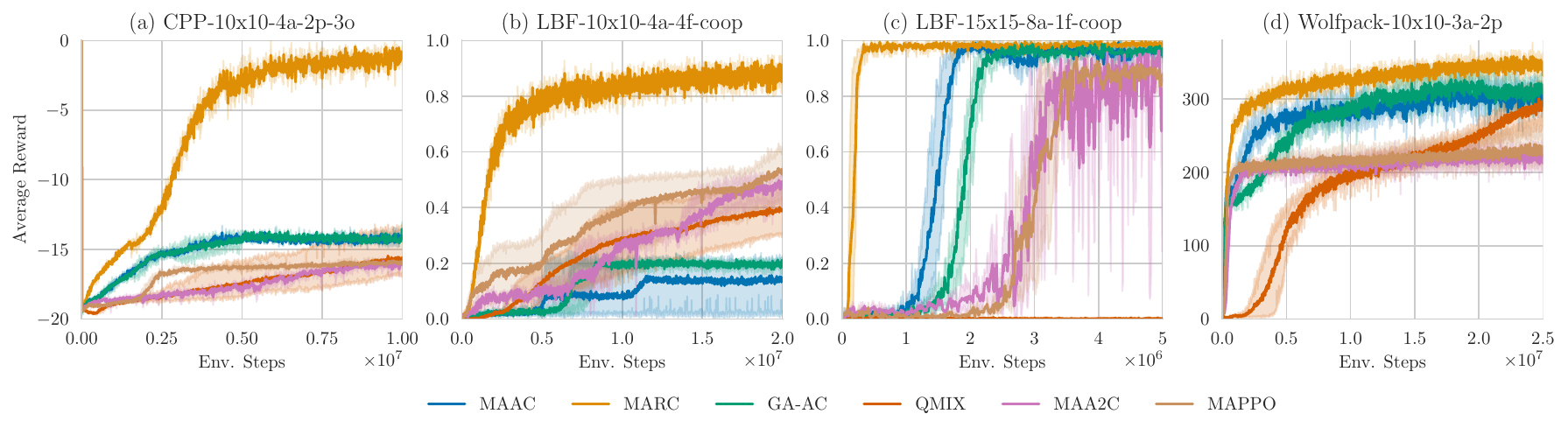}
  \caption{Mean average performance and $95\%$ confidence interval for all discrete tasks. For each model, we run 3 random seeds.} %(\subref{fig:cpp}) presents collaborative pick-and-place, (\subref{fig:lbf-15x15_8a_1f_coop}) and (\subref{fig:lbf-10x10_4a_4f_coop}) are two different settings of the LBF environment whilst (\subref{fig:wolfpack}) is our setup for the wolfpack environment.
  \label{fig:results}
  % \Description{Asymptotic performance  for the LBF environment and wolfpack}
\end{figure*}

\subsection{Asymptotic Performance and Sample Efficiency in Discrete Domains}

In this section, we present a comparative analysis of the asymptotic performance and sample efficiency as illustrated in Figure \ref{fig:results}. Asymptotically, MARC is competitive and outperforms all baselines across the implemented tasks. Additionally, MARC demonstrates superior sample efficiency, learning all the tasks the fastest. In the LBF-15x15-8a-1f-coop task, MARC reaches an average performance of 99\% after 5.9e5 environment steps, whereas the second-best algorithm, MAAC, takes 7.3 times the number of steps to reach the same performance.

The most significant margins in asymptotic performance are achieved in CPP and LBF-10x10-4a-4f-coop, where MARC achieves a performance gain of 69.9\% and 35.2\% respectively, as displayed in Figure 2a and Figure 2b. They require a high level of coordination and spatial understanding between entities to succeed in the task. In the LBF-10x10-4a-4f-coop setting, MARC reaches 26\% of the maximum returns, on average, in 1e6 steps, while the second-best algorithm, MAPPO, reaches the same performance in 5.6 times the number of steps. MAAC performs relatively well in tasks that highly depend on coordination between agents, such as LBF-15x15-8a-1f-coop and Wolfpack, as visualized in Figure 2c and Figure 2d, respectively. However, MAAC's performance deteriorates in CPP and LBF-10x10-4a-4f-coop, where information about objects is essential to gain a good understanding of the environment. GA-AC and MAAC do not have a significant performance difference, indicating that the additional hard-attention layer on the agent interactions does not dramatically impact performance in spatially demanding tasks involving reasoning over environment objects. MAA2C and MAPPO perform reasonably well in LBF and Wolfpack. As both share the critic and policy network across agents, one hypothesis is that this proves beneficial in highly collaborative and homogeneous tasks. 

Overall, the comparative analysis demonstrates the effectiveness of MARC in achieving superior asymptotic performance and sample efficiency in the selected multi-agent environments. The spatial inductive bias introduced in MARC proves to be beneficial in understanding the relationships between agents and objects, leading to faster learning and better asymptotic performance compared to the baselines. % algorithms. 

\subsection{Generalisation Performance} 
To assess the ability of our method to generalize to out-of-distribution settings, we evaluate our model trained on the most difficult scenario of LBF, 10x10-4a-4f-coop, where MARC achieves $81\%$ of the maximum performance, on a varying number of agents and fruits. We then compare our algorithm by training the best-performing algorithm on this task, MAPPO, with the same varied number of fruits and agents. When reducing the number of agents available to collect fruits to 3, MARC still achieves $38\%$ of the performance, whilst MAPPO's performance fully deteriorates to $0\%$. Increasing the number of agents by 1 makes the task easier and yields an improved performance of $93\%$ vs. $88\%$ for MAPPO. This indicates that MARC learns an invariance to the number of agents. The performance decreases to 59\% with an increase in fruits (from 4 to 6 fruits), but given that the number of environment steps remains fixed it generally becomes more difficult to fulfill in time and can still be considered robust. In comparison, MAPPO's performance decreases by 40\% down to 19\%. An overview of all generalization results can be found in the appendix.

\subsection{Extension to Continuous Domain}
\begin{table}[htbp]
    \centering
    \begin{tabular}{lcc} % Left-aligned, with fixed columns
        \toprule
        Algorithm & \makecell{3 Agents \\ ($2\times10^6$ steps)} & \makecell{7 Agents \\ ($4\times10^6$ steps) } \\
        % & \makecell{Sparse Rewards\\ ($4\times10^6$)} \\
        \midrule
        % MARC (5 relations)  & $208.9 \pm 8.5$ & $432.5 \pm 4.3$   \\
        MARC   & $212.7 \pm 5.7$ & $468.2 \pm 4.2$   \\
        InforMARL & $193.5 \pm 4.3$ & $426.1 \pm 81.2$\\
        MAAC    & $236.1 \pm 2.9$ & $527.9 \pm 5.4$ \\
        GA-AC  & $\mathbf{236.6 \pm 3.5}$  & $ \mathbf{530.6 \pm 3.4} $  \\
        MAA2C   & $233.5 \pm 2.1 $& $ 68.8 \pm 393.4 $  \\
        MAPPO   & $109.0 \pm 16.2$ & $304.7 \pm 6.0$\\
        QMIX    & $21.5 \pm 14.8 $ & $-90.0 \pm 79.6$ \\
        \bottomrule
    \end{tabular}
    \caption{Asymptotic performance and standard deviation for the Target task, averaged across 3 seeds.}
    \label{tab:continuous_table}
\end{table}

As seen in Table~\ref{tab:continuous_table}, MARC performs stronger than the SOTA graph-based algorithm InforMARL, underlining the strength of our graph design also in continuous domains. It is also competitive with the best-performing baselines, MAAC, GA-AC and MAA2C. Deeper analysis shows that the performance margin comes from MARC taking, on average, 1-2 steps longer to reach the target. There is a trade-off in abstraction between a simple and efficient abstraction and removing too much information. For example, as environment objects also have velocity in the target task, the agents ideally have a more fine-grained understanding of the proximity to other objects, rather than just knowing once they are adjacent. This could lead to collisions that cannot be avoided or initially going passed their target due to their accumulated velocity. Further experiments, which can be found in the appendix, have confirmed this trade-off, indicating that a coarser abstraction yields improvements in sample efficiency with a decrease in asymptotic performance. Nevertheless, our algorithm demonstrates competitive and robust performance even for the continuous case.

%In this case, it seems that the loss in granularity of the state representation results in less targeted movement.

\section{Ablation Studies}
\label{sec:ablation}

Given the immense flexibility of graph architectures, we aim to shed light on how different design choices affect performance by systematically varying the following aspects: 
% this is all measured after 8e6 steps

\subsubsection{Choice of Relations} To understand how the choice of relations impacts performance, we evaluate our experiments with 3 different groups: our \textit{default relations}, \textit{local relations} representing a convolutional kernel, and \textit{all relations} as the union of the two, detailed in the appendix. We found that purely local relations are not sufficient to learn the task, achieving only 10.6\% of the performance achieved by the chosen architecture. This seems intuitive, as the agent gains a deeper spatial understanding if they can infer information from all entities, even if they are further away. Additionally, adding local relations to our default set does not elevate the performance, indicating that our default relations offer a sufficient and strong enough spatial bias.

\subsubsection{Number of Entities} We compared our approach of considering only agents and objects in the graph to using all grid elements as entities. Learning over the full grid compared to our choice of compact representation is, despite being more informative, not as sample efficient, reaching only 46.9\% of the performance achieved by the chosen architecture in $8e6$ environment steps, along with a higher computational cost.

\subsubsection{Choice of Graph Architecture} We explore alternative choices of aggregating information from connecting entities in the graph. Whilst the choice is vast, we focus on previous work in the single-agent literature, where relational inductive bias between the entities is introduced via multi-head attention \cite{zambaldi2018}. For this, we construct a binary graph and pass it through a Graph Attention Network (GAT) \citet{GAT2018}. Furthermore, we combine the approaches of spatial relations and varying importance between entities as in GATs by using an R-GAT layer \cite{RGAT2019} on the graph constructed in Section~\ref{sec:method}. For a detailed display of these alternative implementations, we refer to the appendix. 
Our R-GAT and R-GCN implementation yield indistinguishable performance, indicating that implicitly specifying different importance between entities does not yield a more expressive representation and its computation is therefore not required. In contrast, the use of a GAT layer yields suboptimal performance, asymptotically reaching only 23.4\% of the chosen architecture's performance. The non-spatial, weighted interactions among entities might not serve as a robust inductive bias to effectively reason about the inherent structure of the task.
% We note that this implementation does not constitute a state abstraction as we maintain all available information in the graph.

%Ultimately, the observation encoder is, in principle, agnostic to the backbone MARL algorithm itself. We therefore test this hypothesis by combining the relational observation encoder with MAPPO as strong and popular alternative algorithm. We compare our implementation with the original MAPPO implementation, setting the hyperparameters to be the same for a fair comparison. Our results show that the addition of the relational encoder improves sample efficiency and asymptotic performance for the considered task.

\section{Conclusion and Future Work}

In this work, we presented a relational state abstraction approach for MARL and demonstrated its effectiveness in environments requiring spatial reasoning and coordination among agents. By incorporating spatial inductive biases into our abstraction, we achieved significant improvements in sample efficiency and asymptotic performance compared to SOTA MARL algorithms. Our findings provide strong evidence for the potential of leveraging relational inductive biases to address the challenges of sample efficiency and generalization in MARL. 

To further enhance our method, future research could explore the incorporation of inductive biases beyond spatial reasoning, an even stronger incorporation of structured representations, for example into the policy network as well, and the fine-tuning to more complex, high-dimensional environments. Investigating the interpretability and transparency of the structured representation could also facilitate the deployment into real-world scenarios.

\section*{Acknowledgments}
This work was supported by the UK Engineering and Physical Sciences Research Council (EPSRC EP/W523793/1), through the Statistics Centre for Doctoral Training at the University of Warwick and from a UKRI Turing AI Acceleration Fellowship (EPSRC EP/V024868/1).

\bibliography{aaai25}

\begin{thebibliography}{55}
\providecommand{\natexlab}[1]{#1}

\bibitem[{Abdel-Aziz et~al.(2024)Abdel-Aziz, Elbamby, Samarakoon, and Bennis}]{Abdel2024}
Abdel-Aziz, M.~K.; Elbamby, M.~S.; Samarakoon, S.; and Bennis, M. 2024.
\newblock Cooperative Multi-Agent Learning for Navigation via Structured State Abstraction.
\newblock In \emph{IEEE Transactions on Communications}.

\bibitem[{Abel(2022)}]{abel2022theoryabstractionreinforcementlearning}
Abel, D. 2022.
\newblock \emph{A Theory of Abstraction in Reinforcement Learning}.
\newblock Ph.D. thesis, Brown University.

\bibitem[{Agarwal et~al.(2020)Agarwal, Kumar, Sycara, and Lewis}]{Argarwal2020}
Agarwal, A.; Kumar, S.; Sycara, K.; and Lewis, M. 2020.
\newblock Learning Transferable Cooperative Behavior in Multi-Agent Teams.
\newblock In \emph{Proceedings of the 19th International Conference on Autonomous Agents and MultiAgent Systems}, 1741–1743.

\bibitem[{Agogino and Tumer(2012)}]{Airtraffic2010}
Agogino, A.~K.; and Tumer, K. 2012.
\newblock A multiagent approach to managing air traffic flow.
\newblock \emph{Autonomous Agents and Multi-Agent Systems}, 24: 1--25.

\bibitem[{Bapst et~al.(2019)Bapst, Sanchez-Gonzalez, Doersch, Stachenfeld, Kohli, Battaglia, and Hamrick}]{Bapst2019}
Bapst, V.; Sanchez-Gonzalez, A.; Doersch, C.; Stachenfeld, K.~L.; Kohli, P.; Battaglia, P.~W.; and Hamrick, J.~B. 2019.
\newblock Structured agents for physical construction.
\newblock In \emph{Proceedings of the 36th International Conference on Machine Learning}.

\bibitem[{Battaglia et~al.(2018)Battaglia, Hamrick, Bapst, Sanchez, Zambaldi, Malinowski, Tacchetti, Raposo, Santoro, Faulkner, Gulcehre, Song, Ballard, Gilmer, Dahl, Vaswani, Allen, Nash, Langston, Dyer, Heess, Wierstra, Kohli, Botvinick, Vinyals, Li, and Pascanu}]{inductivebias}
Battaglia, P.; Hamrick, J. B.~C.; Bapst, V.; Sanchez, A.; Zambaldi, V.; Malinowski, M.; Tacchetti, A.; Raposo, D.; Santoro, A.; Faulkner, R.; Gulcehre, C.; Song, F.; Ballard, A.; Gilmer, J.; Dahl, G.~E.; Vaswani, A.; Allen, K.; Nash, C.; Langston, V.~J.; Dyer, C.; Heess, N.; Wierstra, D.; Kohli, P.; Botvinick, M.; Vinyals, O.; Li, Y.; and Pascanu, R. 2018.
\newblock Relational inductive biases, deep learning, and graph networks.
\newblock arXiv:1806.01261.

\bibitem[{Bengio, Courville, and Vincent(2012)}]{Bengio2012}
Bengio, Y.; Courville, A.~C.; and Vincent, P. 2012.
\newblock Representation Learning: A Review and New Perspectives.
\newblock \emph{IEEE Transactions on Pattern Analysis and Machine Intelligence}, 35: 1798--1828.

\bibitem[{Busbridge et~al.(2019)Busbridge, Sherburn, Cavallo, and Hammerla}]{RGAT2019}
Busbridge, D.; Sherburn, D.; Cavallo, P.; and Hammerla, N.~Y. 2019.
\newblock Relational Graph Attention Networks.
\newblock arXiv:1904.05811.

\bibitem[{Christianos, Schäfer, and Albrecht(2020)}]{lbf2020}
Christianos, F.; Schäfer, L.; and Albrecht, S.~V. 2020.
\newblock Shared Experience Actor-Critic for Multi-Agent Reinforcement Learning.
\newblock In \emph{Advances in Neural Information Processing Systems}, 10707--10717.

\bibitem[{Croonenborghs et~al.(2006)Croonenborghs, Tuyls, Ramon, and Bruynooghe}]{RMARL2006}
Croonenborghs, T.; Tuyls, K.; Ramon, J.; and Bruynooghe, M. 2006.
\newblock Multi-agent Relational Reinforcement Learning.
\newblock In \emph{Learning and Adaption in Multi-Agent Systems}, 192--206.

\bibitem[{Driessens and D{\v{z}}eroski(2001)}]{driessens2001}
Driessens, K.; and D{\v{z}}eroski, S. 2001.
\newblock Integrating guidance into relational reinforcement learning.
\newblock \emph{Machine Learning}, 116--127.

\bibitem[{D{\v{z}}eroski, De~Raedt, and Driessens(2001)}]{RRL2001}
D{\v{z}}eroski, S.; De~Raedt, L.; and Driessens, K. 2001.
\newblock Relational reinforcement learning.
\newblock \emph{Machine Learning}, 7--52.

\bibitem[{Garnelo, Arulkumaran, and Shanahan(2016)}]{Garnelo2016}
Garnelo, M.; Arulkumaran, K.; and Shanahan, M. 2016.
\newblock Towards Deep Symbolic Reinforcement Learning.
\newblock arXiv:1609.05518.

\bibitem[{Gilmer et~al.(2017)Gilmer, Schoenholz, Riley, Vinyals, and Dahl}]{gilmer2017}
Gilmer, J.; Schoenholz, S.~S.; Riley, P.~F.; Vinyals, O.; and Dahl, G.~E. 2017.
\newblock Neural message passing for quantum chemistry.
\newblock In \emph{Proceedings of the 34th International Conference on Machine Learning}, 1263--1272.

\bibitem[{Haarnoja et~al.(2018)Haarnoja, Zhou, Abbeel, and Levine}]{SAC}
Haarnoja, T.; Zhou, A.; Abbeel, P.; and Levine, S. 2018.
\newblock Soft Actor-Critic: Off-Policy Maximum Entropy Deep Reinforcement Learning with a Stochastic Actor.
\newblock In \emph{Proceedings of the 35th International Conference on Machine Learning}, 1861--1870.

\bibitem[{Hamilton, Ying, and Leskovec(2017)}]{hamilton2017inductive}
Hamilton, W.; Ying, Z.; and Leskovec, J. 2017.
\newblock Inductive representation learning on large graphs.
\newblock \emph{Advances in neural information processing systems}, 30.

\bibitem[{Iqbal and Sha(2019)}]{MAAC}
Iqbal, S.; and Sha, F. 2019.
\newblock Actor-Attention-Critic for Multi-Agent Reinforcement Learning.
\newblock In \emph{Proceedings of the 36th International Conference on Machine Learning}, 2961--2970.

\bibitem[{Jiang et~al.(2020)Jiang, Dun, Huang, and Lu}]{DGN2020}
Jiang, J.; Dun, C.; Huang, T.; and Lu, Z. 2020.
\newblock Graph Convolutional Reinforcement Learning.
\newblock In \emph{International Conference on Learning Representations}.

\bibitem[{Jiang et~al.(2021)Jiang, Minervini, Jiang, and Rockt\"{a}schel}]{GTG2021}
Jiang, Z.; Minervini, P.; Jiang, M.; and Rockt\"{a}schel, T. 2021.
\newblock Grid-to-Graph: Flexible Spatial Relational Inductive Biases for Reinforcement Learning.
\newblock In \emph{Proceedings of the 20th International Conference on Autonomous Agents and MultiAgent Systems}, 674–682.

\bibitem[{Khan et~al.(2019)Khan, Tolstaya, Ribeiro, and Kumar}]{GPG2019}
Khan, A.; Tolstaya, E.~V.; Ribeiro, A.; and Kumar, V.~R. 2019.
\newblock Graph Policy Gradients for Large Scale Robot Control.
\newblock In \emph{Conference on Robot Learning}, 823--834.

\bibitem[{Kingma and Ba(2015)}]{Adam}
Kingma, D.~P.; and Ba, J. 2015.
\newblock Adam: {A} Method for Stochastic Optimization.
\newblock In \emph{3rd International Conference on Learning Representations, {ICLR} 2015, San Diego, CA, USA, May 7-9, 2015, Conference Track Proceedings}.

\bibitem[{Kipf et~al.(2018)Kipf, Fetaya, Wang, Welling, and Zemel}]{NRI2018}
Kipf, T.; Fetaya, E.; Wang, K.-C.; Welling, M.; and Zemel, R. 2018.
\newblock Neural Relational Inference for Interacting Systems.
\newblock In \emph{Proceedings of the 35th International Conference on Machine Learning}, 2688--2697.

\bibitem[{Kipf, van~der Pol, and Welling(2020)}]{Kipf2020}
Kipf, T.; van~der Pol, E.; and Welling, M. 2020.
\newblock Contrastive Learning of Structured World Models.
\newblock In \emph{International Conference on Learning Representations}.

\bibitem[{Li et~al.(2022)Li, Xiao, Zhang, Liu, and Shen}]{DILP2022}
Li, G.; Xiao, G.; Zhang, J.; Liu, J.; and Shen, Y. 2022.
\newblock Towards Relational Multi-Agent Reinforcement Learning via Inductive Logic Programming.
\newblock In \emph{Artificial Neural Networks and Machine Learning}, 99--110.

\bibitem[{Li et~al.(2020)Li, Yang, Tomizuka, and Choi}]{Evolvegraph2020}
Li, J.; Yang, F.; Tomizuka, M.; and Choi, C. 2020.
\newblock EvolveGraph: Multi-Agent Trajectory Prediction with Dynamic Relational Reasoning.
\newblock In \emph{Advances in Neural Information Processing Systems}, 19783--19794.

\bibitem[{Li, Walsh, and Littman(2006)}]{li2006towards}
Li, L.; Walsh, T.; and Littman, M. 2006.
\newblock Towards a Unified Theory of State Abstraction for MDPs.
\newblock In \emph{Proceedings of the Ninth International Symposium on Artificial Intelligence and Mathematics}.

\bibitem[{Liu et~al.(2021)Liu, Ren, Yeh, and Schwing}]{VMARL}
Liu, I.-J.; Ren, Z.; Yeh, R.~A.; and Schwing, A.~G. 2021.
\newblock Semantic tracklets: An object-centric representation for visual multi-agent reinforcement learning.
\newblock In \emph{International Conference on Intelligent Robots and Systems}, 5603--5610.

\bibitem[{Liu et~al.(2020)Liu, Wang, Hu, Hao, Chen, and Gao}]{g2anet}
Liu, Y.; Wang, W.; Hu, Y.; Hao, J.; Chen, X.; and Gao, Y. 2020.
\newblock Multi-agent game abstraction via graph attention neural network.
\newblock In \emph{Proceedings of the AAAI conference on artificial intelligence}, 7211--7218.

\bibitem[{Lowe et~al.(2017)Lowe, Wu, Tamar, Harb, Abbeel, and Mordatch}]{MADDPG}
Lowe, R.; Wu, Y.; Tamar, A.; Harb, J.; Abbeel, P.; and Mordatch, I. 2017.
\newblock Multi-Agent Actor-Critic for Mixed Cooperative-Competitive Environments.
\newblock In \emph{Advances in Neural Information Processing Systems}.

\bibitem[{Malysheva, Kudenko, and Shpilman(2019)}]{MAGNET19}
Malysheva, A.; Kudenko, D.; and Shpilman, A. 2019.
\newblock MAGNet: Multi-agent Graph Network for Deep Multi-agent Reinforcement Learning.
\newblock In \emph{XVI International Symposium "Problems of Redundancy in Information and Control Systems" (REDUNDANCY)}, 171--176.

\bibitem[{Matignon, Jeanpierre, and Mouaddib(2012)}]{Multi-robot2012}
Matignon, L.; Jeanpierre, L.; and Mouaddib, A.-I. 2012.
\newblock Coordinated Multi-Robot Exploration Under Communication Constraints Using Decentralized Markov Decision Processes.
\newblock In \emph{Proceedings of the AAAI Conference on Artificial Intelligence}, 2017--2023.

\bibitem[{Mnih et~al.(2015)Mnih, Kavukcuoglu, Silver, Rusu, Veness, Bellemare, Graves, Riedmiller, Fidjeland, Ostrovski, Petersen, Beattie, Sadik, Antonoglou, King, Kumaran, Wierstra, Legg, and Hassabis}]{Mnih2015}
Mnih, V.; Kavukcuoglu, K.; Silver, D.; Rusu, A.~A.; Veness, J.; Bellemare, M.~G.; Graves, A.; Riedmiller, M.~A.; Fidjeland, A.; Ostrovski, G.; Petersen, S.; Beattie, C.; Sadik, A.; Antonoglou, I.; King, H.; Kumaran, D.; Wierstra, D.; Legg, S.; and Hassabis, D. 2015.
\newblock Human-level control through deep reinforcement learning.
\newblock \emph{Nature}, 518: 529--533.

\bibitem[{Mohan, Zhang, and Lindauer(2024)}]{StructureRL}
Mohan, A.; Zhang, A.; and Lindauer, M. 2024.
\newblock Structure in Deep Reinforcement Learning: A Survey and Open Problems.
\newblock \emph{Journal of Artificial Intelligence Research}, 79.

\bibitem[{Nayak et~al.(2023)Nayak, Choi, Ding, Dolan, Gopalakrishnan, and Balakrishnan}]{informal}
Nayak, S.; Choi, K.; Ding, W.; Dolan, S.; Gopalakrishnan, K.; and Balakrishnan, H. 2023.
\newblock Scalable Multi-Agent Reinforcement Learning through Intelligent Information Aggregation.
\newblock In \emph{Proceedings of the 40th International Conference on Machine Learning}, 25817--25833.

\bibitem[{Niu, Paleja, and Gombolay(2021)}]{MAGIC}
Niu, Y.; Paleja, R.; and Gombolay, M. 2021.
\newblock Multi-Agent Graph-Attention Communication and Teaming.
\newblock In \emph{Proceedings of the 20th International Conference on Autonomous Agents and MultiAgent Systems}, 964–973.

\bibitem[{Papoudakis et~al.(2020)Papoudakis, Christianos, Sch{\"a}fer, and Albrecht}]{papoudakis2020benchmarking}
Papoudakis, G.; Christianos, F.; Sch{\"a}fer, L.; and Albrecht, S.~V. 2020.
\newblock Benchmarking multi-agent deep reinforcement learning algorithms in cooperative tasks.
\newblock In \emph{NeurIPS Datasets and Benchmarks}.

\bibitem[{Ponsen et~al.(2010)Ponsen, Croonenborghs, Tuyls, Ramon, Driessens, Herik, and Postma}]{CommRMARL2010}
Ponsen, M.; Croonenborghs, T.; Tuyls, K.; Ramon, J.; Driessens, K.; Herik, H.; and Postma, E. 2010.
\newblock \emph{Learning with Whom to Communicate Using Relational Reinforcement Learning}, 45--63.
\newblock Studies in Computational Intelligence. Springer.

\bibitem[{Rahman et~al.(2023)Rahman, Carlucho, H{\"o}pner, and Albrecht}]{GPL2023}
Rahman, A.; Carlucho, I.; H{\"o}pner, N.; and Albrecht, S.~V. 2023.
\newblock A general learning framework for open ad hoc teamwork using graph-based policy learning.
\newblock \emph{Journal of Machine Learning Research}, 24: 1--74.

\bibitem[{Rashid et~al.(2018)Rashid, Samvelyan, Witt, Farquhar, Foerster, and Whiteson}]{QMIX}
Rashid, T.; Samvelyan, M.; Witt, C. S.~D.; Farquhar, G.; Foerster, J.~N.; and Whiteson, S. 2018.
\newblock QMIX: Monotonic Value Function Factorisation for Deep Multi-Agent Reinforcement Learning.
\newblock In \emph{Proceedings of the 35th International Conference on Machine Learning}.

\bibitem[{Samet(1984)}]{quadtree}
Samet, H. 1984.
\newblock The Quadtree and Related Hierarchical Data Structures.
\newblock \emph{ACM Computing Surveys}, 16: 187–260.

\bibitem[{Sanner and Boutilier(2009)}]{sanner2009}
Sanner, S.; and Boutilier, C. 2009.
\newblock Practical solution techniques for first-order MDPs.
\newblock \emph{Artificial Intelligence}, 173: 748--788.

\bibitem[{Scarselli et~al.(2009)Scarselli, Gori, Tsoi, Hagenbuchner, and Monfardini}]{Scarselli2009}
Scarselli, F.; Gori, M.; Tsoi, A.~C.; Hagenbuchner, M.; and Monfardini, G. 2009.
\newblock The Graph Neural Network Model.
\newblock \emph{IEEE Transactions on Neural Networks}, 20: 61--80.

\bibitem[{Schlichtkrull et~al.(2018)Schlichtkrull, Kipf, Bloem, van~den Berg, Titov, and Welling}]{Schlichtkrull2018}
Schlichtkrull, M.; Kipf, T.~N.; Bloem, P.; van~den Berg, R.; Titov, I.; and Welling, M. 2018.
\newblock Modeling Relational Data with Graph Convolutional Networks.
\newblock In \emph{The Semantic Web}, 593--607.

\bibitem[{Schulman et~al.(2017)Schulman, Wolski, Dhariwal, Radford, and Klimov}]{PPO}
Schulman, J.; Wolski, F.; Dhariwal, P.; Radford, A.; and Klimov, O. 2017.
\newblock Proximal Policy Optimization Algorithms.
\newblock arXiv:1707.06347.

\bibitem[{Shalev-Shwartz, Shammah, and Shashua(2016)}]{AutoVehicles2016}
Shalev-Shwartz, S.; Shammah, S.; and Shashua, A. 2016.
\newblock Safe, Multi-Agent, Reinforcement Learning for Autonomous Driving.
\newblock arXiv:1610.03295.

\bibitem[{Song, Stojanovic, and Chitre(2019)}]{underwater}
Song, A.; Stojanovic, M.; and Chitre, M. 2019.
\newblock Underwater Acoustic Communications: Where we Stand and What is Next?
\newblock \emph{IEEE Journal of Oceanic Engineering}, 44.

\bibitem[{Tacchetti et~al.(2019)Tacchetti, Song, Mediano, Zambaldi, Kramár, Rabinowitz, Graepel, Botvinick, and Battaglia}]{RFM2018}
Tacchetti, A.; Song, H.~F.; Mediano, P. A.~M.; Zambaldi, V.; Kramár, J.; Rabinowitz, N.~C.; Graepel, T.; Botvinick, M.; and Battaglia, P.~W. 2019.
\newblock Relational Forward Models for Multi-Agent Learning.
\newblock In \emph{International Conference on Learning Representations}.

\bibitem[{Vaswani et~al.(2017)Vaswani, Shazeer, Parmar, Uszkoreit, Jones, Gomez, Kaiser, and Polosukhin}]{Vaswani2017}
Vaswani, A.; Shazeer, N.; Parmar, N.; Uszkoreit, J.; Jones, L.; Gomez, A.~N.; Kaiser, L.~u.; and Polosukhin, I. 2017.
\newblock Attention is All you Need.
\newblock In \emph{Advances in Neural Information Processing Systems}.

\bibitem[{Veličković et~al.(2018)Veličković, Cucurull, Casanova, Romero, Liò, and Bengio}]{GAT2018}
Veličković, P.; Cucurull, G.; Casanova, A.; Romero, A.; Liò, P.; and Bengio, Y. 2018.
\newblock Graph Attention Networks.
\newblock In \emph{International Conference on Learning Representations}.

\bibitem[{Welling and Kipf(2016)}]{Welling2016}
Welling, M.; and Kipf, T.~N. 2016.
\newblock Semi-supervised classification with graph convolutional networks.
\newblock In \emph{International Conference on Learning Representations}.

\bibitem[{Yu et~al.(2022)Yu, Velu, Vinitsky, Gao, Wang, Bayen, and WU}]{MAPPO}
Yu, C.; Velu, A.; Vinitsky, E.; Gao, J.; Wang, Y.; Bayen, A.; and WU, Y. 2022.
\newblock The Surprising Effectiveness of PPO in Cooperative Multi-Agent Games.
\newblock In \emph{Advances in Neural Information Processing Systems}, 24611--24624.

\bibitem[{Zambaldi et~al.(2019)Zambaldi, Raposo, Santoro, Bapst, Li, Babuschkin, Tuyls, Reichert, Lillicrap, Lockhart, Shanahan, Langston, Pascanu, Botvinick, Vinyals, and Battaglia}]{zambaldi2018}
Zambaldi, V.; Raposo, D.; Santoro, A.; Bapst, V.; Li, Y.; Babuschkin, I.; Tuyls, K.; Reichert, D.; Lillicrap, T.; Lockhart, E.; Shanahan, M.; Langston, V.; Pascanu, R.; Botvinick, M.; Vinyals, O.; and Battaglia, P. 2019.
\newblock Deep reinforcement learning with relational inductive biases.
\newblock In \emph{International Conference on Learning Representations}.

\bibitem[{Zhang et~al.(2021{\natexlab{a}})Zhang, McAllister, Calandra, Gal, and Levine}]{Zhang2021}
Zhang, A.; McAllister, R.~T.; Calandra, R.; Gal, Y.; and Levine, S. 2021{\natexlab{a}}.
\newblock Learning Invariant Representations for Reinforcement Learning without Reconstruction.
\newblock In \emph{International Conference on Learning Representations}.

\bibitem[{Zhang et~al.(2021{\natexlab{b}})Zhang, Liu, Xu, Huang, Mao, and Carie}]{sriac2021}
Zhang, X.; Liu, Y.; Xu, X.; Huang, Q.; Mao, H.; and Carie, A. 2021{\natexlab{b}}.
\newblock Structural relational inference actor-critic for multi-agent reinforcement learning.
\newblock \emph{Neurocomputing}, 459: 383--394.

\bibitem[{Zucker(2003)}]{Zucker}
Zucker, J. 2003.
\newblock A grounded theory of Abstraction in Artificial Intelligence.
\newblock \emph{Philosophical transactions of the Royal Society of London. Series B, Biological sciences}, 358: 1293--309.

\end{thebibliography}

\section*{Data Appendix}
\subsection*{Environments}

To explore the effectiveness of learning a relational state representation in MARL, we have selected a diverse set of environments that offer a suitable testbed for examining the learning of spatial relationships between pairs of entities. The chosen environments all involve multiple agents interacting with each other and their surroundings in ways that require them to reason about the relative positions, distances, and spatial configurations of entities in the environment.

\subsubsection*{Collaborative Pick and Place} 
\begin{figure}[h]
    \centering
    \includegraphics[width=0.5\linewidth]{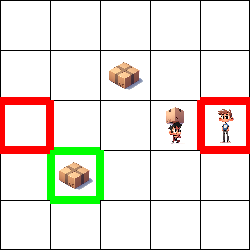}
    \caption{Collaborative pick and place environment on a 5x5 grid with 1 picker agent, 1 delivery agent, 3 boxes and 3 goal locations.}
\label{fig:cpp_display}
\end{figure}

The Collaborative Pick and Place (CPP) environment introduces a novel multi-agent challenge that involves heterogeneous agent roles working together to complete a task. In this environment, picker agents and delivery agents must cooperate to pick up boxes and drop them off at designated goal locations. The picker agents are responsible for collecting the boxes, while the delivery agents take the boxes from the picker agents and deliver them to the goal locations. Once a box is placed at a goal, it secures the spot, preventing any other boxes from being placed there. The agents operate within a grid world and, at the beginning of each episode, the environment is initialized with agents, boxes, and goals randomly distributed across the grid. An example of this environment involving two agents and two goals is illustrated in Figure \ref{fig:cpp_display}. 

Our experiments are conducted in a more complex $10 \times 10$ grid setup that includes 2 picker agents, 2 delivery agents, and 3 boxes. The agents have a set of six actions at their disposal: move up, down, left, right, pass a box, or wait. Rewards are assigned based on successful interactions between the agents and the environment. Picker agents receive rewards for picking up boxes, while both types of agents earn rewards for successful passes and drop-offs. To encourage cooperation and discourage redundancy, the reward structure is designed as follows: the first pass between a picker agent and a delivery agent grants each agent a reward of 0.5, while repeated passes of the same box result in a penalty of -1. To promote efficient task completion, agents receive a step penalty of -0.1 at each time step, incentivizing them to finish the task quickly. Additionally, if the agents complete the task within the 50-step limit per episode, they receive a completion bonus of 1.

The CPP environment offers flexibility in terms of grid size, the number of agents, and the number of boxes, making it adaptable to various experimental setups. This versatility allows researchers to investigate different aspects of multi-agent coordination and task allocation strategies among heterogeneous agents. 

The observations entail information about the position, entity type, agent type, and whether or not an agent is carrying an object of all entities. The code is available at \url{https://github.com/gmontana/CollaborativePickAndPlaceEnv}.

%For the graph-structured observation, the entity features are binary indicator features $$z_v = (\textit{Agent}_v, \textit{Object}_v, \textit{Goal}_v, \textit{Carrying}_v, \textit{Picker}_v)$$ for all $v \in \gV$.

\subsubsection*{Level-based Foraging} 

% \begin{wrapfigure}{R}{.4\linewidth}
% \centering
%     \includegraphics[width=0.4\linewidth]{figures/step_16.pdf}
%     \caption{Level-based Foraging environment with 4 agents and fruits that become trees once picked.}
%     \label{fig:lbf_env}
% \end{wrapfigure}

%An agent can collect fruit to receive a reward only if their level is equal to or higher than the fruit's level. As fruit levels can be larger than the level of individual agents, it also presents a collaborative task. We can enforce such cooperation by setting the level of the fruits equal to the sum of the 3 agents with the highest level. The reward received for foraging depends on the level of the agent, the foraged fruit, and potentially adjacent agents. The LBF environment allows for flexibility in difficulty by changing the number of agents, fruits, grid size and forcing cooperation. 
\begin{figure}[h]
\centering
    \includegraphics[width=0.5\linewidth]{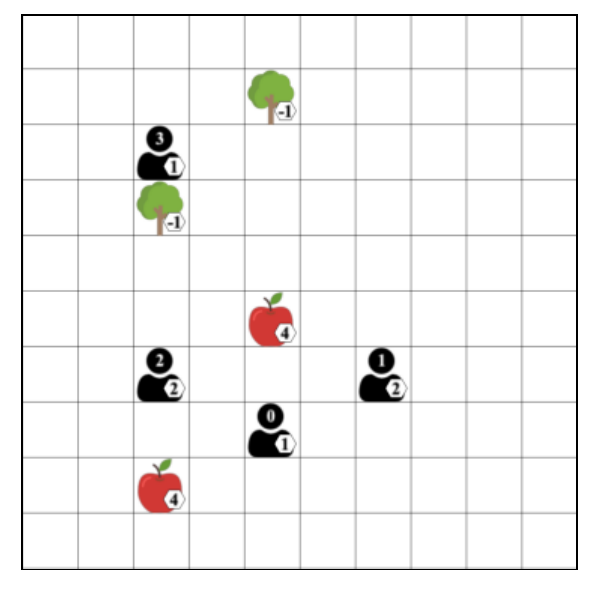}
    \caption{Level-based Foraging environment with 4 agents and fruits that become trees once picked.}
    \label{fig:lbf_env}
\end{figure}
The Level-based Foraging (LBF) environment, originally introduced by \citet{lbf2020}, places agents in a grid world where they are tasked with collecting fruits to receive rewards. The environment incorporates a level-based system that determines an agent's ability to collect a fruit. An agent can only collect a fruit if their level is equal to or higher than the fruit's level. This mechanic introduces a collaborative aspect to the task, as fruit levels can be set higher than the level of individual agents, requiring them to work together. This is a sparse environment, as the agents only receive a reward when they successfully collect a fruit, where the reward is proportional to the agent's level.

% In our implementation, we enforce cooperation by setting the level of the fruits equal to the sum of the levels of the three agents with the highest levels. The reward received for foraging a fruit depends on the level of the agent, the level of the foraged fruit, and potentially the levels of adjacent agents. The LBF environment offers flexibility in adjusting the difficulty by modifying the number of agents, fruits, grid size, and the degree of enforced cooperation.

We have made a notable alteration to the original LBF environment as seen in  Figure~\ref{fig:lbf_env} by introducing the concept of trees. In our version, fruits are assumed to be on trees that remain on the grid even after the fruit has been collected. These trees have a value of $-1$ and serve as obstacles that the agents must navigate around. This modification adds a layer of complexity to the task, requiring agents to possess higher relational reasoning capabilities to recognize and avoid the noncollectable tree obstacles while searching for fruits \footnote{Our fork including our modifications for this environment can be found under https://github.com:sharlinu/lb-foraging}. 

To evaluate the agents' ability to cooperate under challenging conditions, we conduct experiments on a $10\times10$ grid with 4 agents and 4 fruits, enforcing cooperation (denoted as 10x10-4a-4f-coop). This setup presents a scenario with sparse rewards, demanding effective coordination among the agents to succeed. Furthermore, to assess the scalability of the agents' strategies, we extend the environment to a larger $15 \times 15$ grid with 8 agents and 1 fruit, still enforcing cooperation (denoted as 15x15-8a-1f-coop). This expanded setup tests the agents' ability to coordinate and adapt their strategies when working with a larger number of agents in a more complex environment.

The observations entail information about the position, entity type and level of all entities.

%For the graph-structured observation, the entity features are binary indicator features $$z_v = (\textit{Agent}_v, \textit{Agent Identity}_v, \textit{Object}_v, \textit{Goal}_v, \textit{Carrying}_v, \textit{Picker}_v)$$ for all $v \in \gV$.

\subsubsection*{Wolfpack} 

\begin{figure}[h]
\centering
    \includegraphics[width=0.5\linewidth]{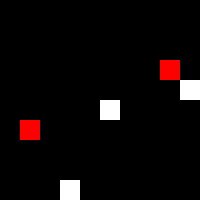}
    \caption{Wolfpack environment with 3 predator agents coordinating to catch 2 moving prey targets.}
    \label{fig:Wolfpack_env}
\end{figure}

Wolfpack is a MARL environment inspired by the implementation of \cite{GPL2023}. In this environment, a team of predator agents is tasked with capturing prey within a 2D grid world. The predators must learn to coordinate their actions and form packs to successfully surround and capture the prey.
The objective of the predator agents is to capture the prey as efficiently as possible. To capture prey, at least two predator agents must surround it by occupying adjacent grid cells. When a prey is successfully captured, the predator agents involved in the capture are rewarded based on the size of the pack. The captured prey is then removed from the grid and respawned at a random location. 

In our specific implementation, we place 3 predator agents and 2 prey in a $10\times10$ grid. The predator agents have full observability, meaning they can see the positions of all objects within the grid. However, the prey agents, which are trained using Deep Q-Networks (DQN) \cite{Mnih2015}, operate under partial observability and can only perceive a 3x3 grid centered on their position. The predator agents are allowed to move in any direction (up, down, left, right) or choose to remain stationary at each time step. The prey agents, on the other hand, follow their own learned policy based on DQN.

In a departure from the original setup, we have modified the reward structure to introduce sparse rewards. We have removed the additional rewards based on the proximity of predator agents to the prey, which were present in the original implementation. This change significantly weakens the learning signal, making the task more challenging for the predator agents to learn optimal coordination strategies \footnote{Our fork including our modifications for this environment can be found under https://github.com:sharlinu/wolfpack}.

The observations entail information about the position and agent type of all entities.

\subsubsection*{Target}

\begin{figure}[h]
\centering
    \includegraphics[width=0.5\linewidth]{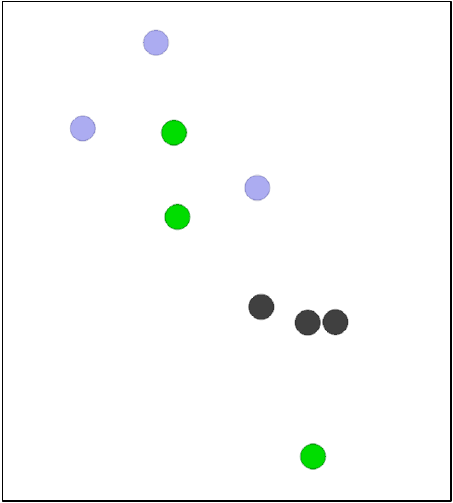}
    \caption{Navigation task with 3 agents aiming to reach target landmarks whilst avoiding obscuring obstacles.}
    \label{fig:mpe_env}
\end{figure}

In the target task, agents try to minimize the distance to specific target landmarks while navigating through moving obstacles and other agents. The environment rewards efficient pathfinding and penalizes collisions, forcing agents to balance speed and caution. This setup requires agents to handle various movements and interactions, similar to real-world scenarios. In our experiment, we test this setting with 3 and 7 agents, that need to reach their assigned markets (3 and 7 targets respectively) whilst avoiding collision with 3 obstacles and the other agents. The observations entail information about the position, velocity and entity type of all entities.

\subsection*{Additional Plots}

\begin{figure*}[t]
    \centering
    \includegraphics[width=\linewidth]{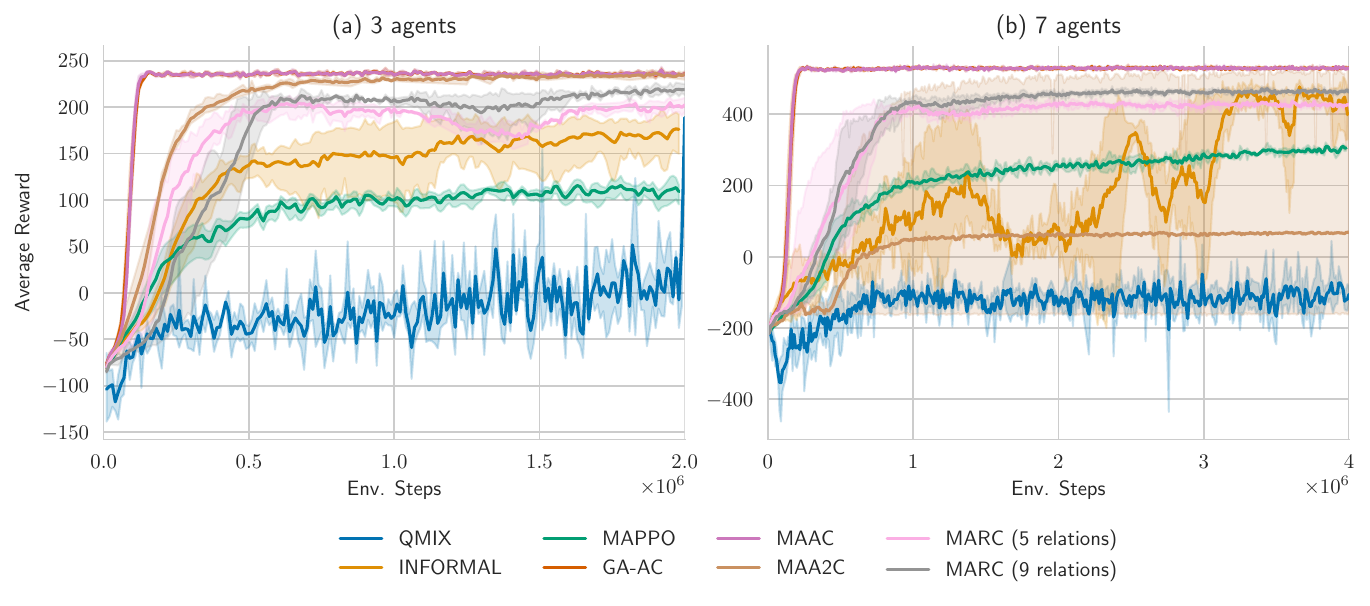}
    \caption{Mean average performance and $95\%$ confidence interval for the continuous target tasks. For each model, we run 3 random seeds.} %
\label{fig:continuous_results}
\end{figure*}

\begin{figure}[h]
    \centering
    \includegraphics[width=\linewidth]{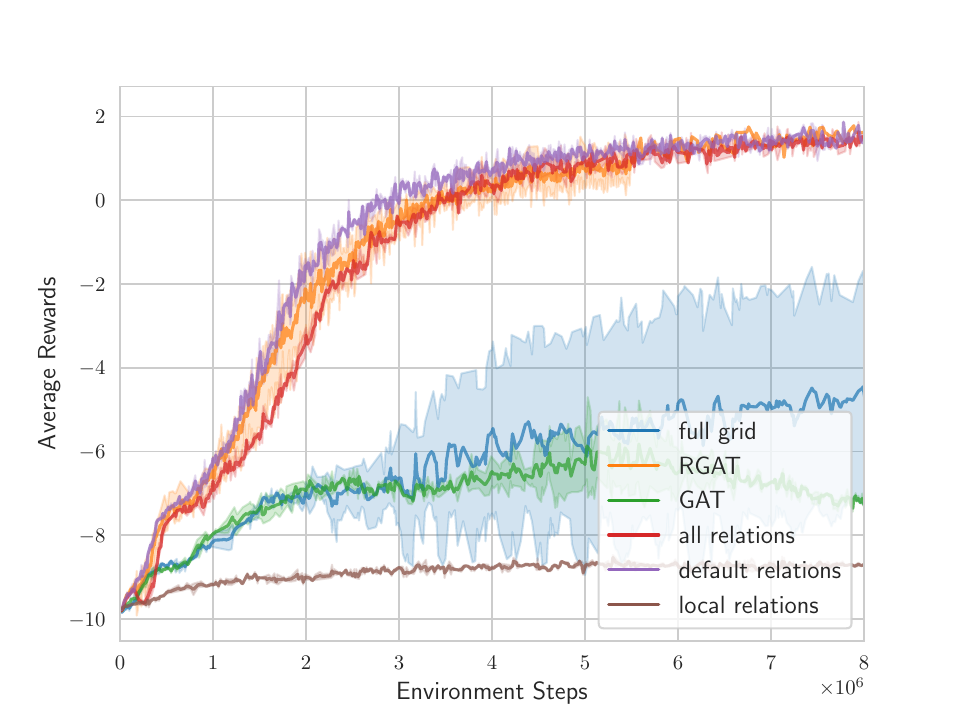}
    \caption{Training curves for a 10x10 CPP environment with 1 picker agent, 1 dropper agent and 2 objects, showing performance for varying graph architectures, sets of relations and number of entities, averaged across 2 seeds.}
    \label{fig: ablation}
\end{figure}

We provide the performance curves for the continuous tasks in Figure~\ref{fig:continuous_results} and for the ablation studies in Figure~\ref{fig: ablation}.

\section*{Technical Appendix}

\subsection*{Constructing Edges for the Observation Graph} \label{sec:relations}

\begin{figure*}
\centering
\begin{subfigure}[t]{.24\linewidth}
    \centering
    \includegraphics[width=\linewidth]{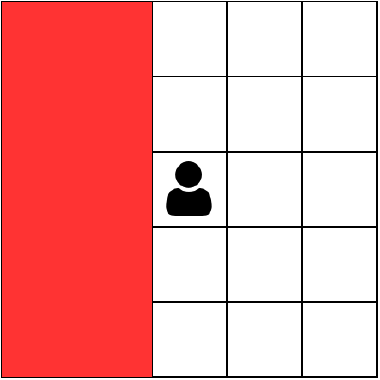}
    \caption{left}
\end{subfigure}
\begin{subfigure}[t]{.24\linewidth}
    \centering
    \includegraphics[width=\linewidth]{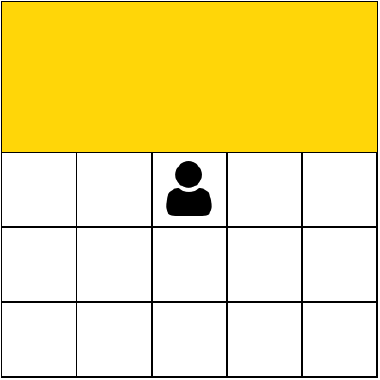}
    \caption{top}
\end{subfigure}
\begin{subfigure}[t]{.24\linewidth}
    \centering
    \includegraphics[width=\linewidth]{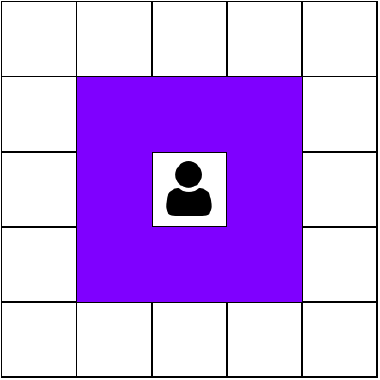}
    \caption{adjacent}
\end{subfigure}
\begin{subfigure}[t]{.24\linewidth}
    \centering
    \includegraphics[width=\linewidth]{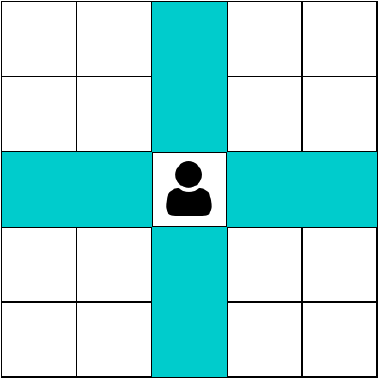}
    \caption{aligned}
\end{subfigure}
\caption{Examples of the spatial determination rules employed in our methodology: given an entity and its position on a grid, the colored areas represent the areas under which a specific relational rule would hold if another entity is positioned there. }
\label{fig:relational_rules}
\end{figure*}

To model interactions and proximities in the observation graph, we define relationships between entities based on their spatial arrangements. These relationships are categorized into three distinct groups: Remote Relations, Contiguous Relations, and Local Relations. Each group serves a specific purpose and represents different levels of proximity and interaction potential between entities.

\subsubsection*{Remote Relations}
Remote relations identify long-range interactions where entities do not need to be immediately adjacent:
\begin{align*}
\text{left}(a,b) &\leftarrow x_a < x_b, \\
\text{right}(a,b) &\leftarrow x_a > x_b,\\
\text{down}(a,b) &\leftarrow y_a < y_b, \\
\text{top}(a,b) &\leftarrow y_a > y_b, 
\end{align*}

\subsubsection*{Contiguous Relations}
Contiguous relations define direct adjacency or alignment, suitable for modeling interactions within immediate reach:
\begin{align*}
\text{aligned}(a,b) &\leftarrow (x_a = x_b) \wedge (y_a = y_b),  \\
\text{adjacent}(a,b) &\leftarrow (|x_a - x_b| \leq 1 ) \vee (|y_a - y_b| \leq 1), 
\end{align*}

\subsubsection*{Local Relations}
Local relations are more granular, detailing the specific neighboring positions around an entity, and are crucial for detailed spatial reasoning:
\begin{align*}
\text{rightAdj}(a,b) &\leftarrow (x_a = x_b + 1) \wedge (y_a = y_b),  \\
\text{leftAdj}(a,b) &\leftarrow (x_a = x_b - 1) \wedge (y_a = y_b), \\
\text{topAdj}(a,b) &\leftarrow (x_a = x_b) \wedge (y_a = y_b + 1),  \\
\text{bottomAdj}(a,b) &\leftarrow (x_a = x_b) \wedge (y_a = y_b - 1),  \\
\text{bottomLeftAdj}(a,b) &\leftarrow (x_a = x_b - 1) \wedge (y_a = y_b -1 ), \\
\text{bottomRightAdj}(a,b) &\leftarrow (x_a = x_b + 1) \wedge (y_a = y_b -1 ),  \\
\text{topLeftAdj}(a,b) &\leftarrow (x_a = x_b - 1) \wedge (y_a = y_b + 1 ), \\
\text{topRightAdj}(a,b) &\leftarrow (x_a = x_b + 1) \wedge (y_a = y_b + 1), 
\end{align*}

An illustrative example of a few of these relations can be seen in Figure~\ref{fig:relational_rules}. For our ablation studies, we categorize the relations into specific groups based on their use:
\begin{itemize}
    \item \textbf{Default set of relations:} These include the most commonly used spatial relationships which cover basic proximity and directional interactions. The set comprises:
    \[
    \{\text{adjacent}, \text{aligned}, \text{left}, \text{right}, \text{top}, \text{bottom}\}
    \]
    
    \item \textbf{Local set of relations:} This set includes more detailed and localized spatial relations, providing finer control and specificity for modeling interactions:
    \begin{align*}
    \{&\text{leftAdj}, \text{rightAdj}, \text{topAdj}, \text{topLeftAdj}, \text{topRightAdj}, \\    
    &\text{bottomAdj}, \text{bottomLeftAdj}, \text{bottomRightAdj} \}
    \end{align*}
    
    \item \textbf{Set of all relations:} Combines both default and local relations for comprehensive coverage:
    \[
    \{\text{default relations} \cup \text{local relations}\}
    \]
\end{itemize}

Default relations are applied in the discrete task unless specified otherwise, offering a balance between computational efficiency and the resolution of spatial relationships.

\subsubsection{Relations for Continuous Domain}
\begin{figure}[h]
    \centering
    \includegraphics[width=0.5\linewidth]{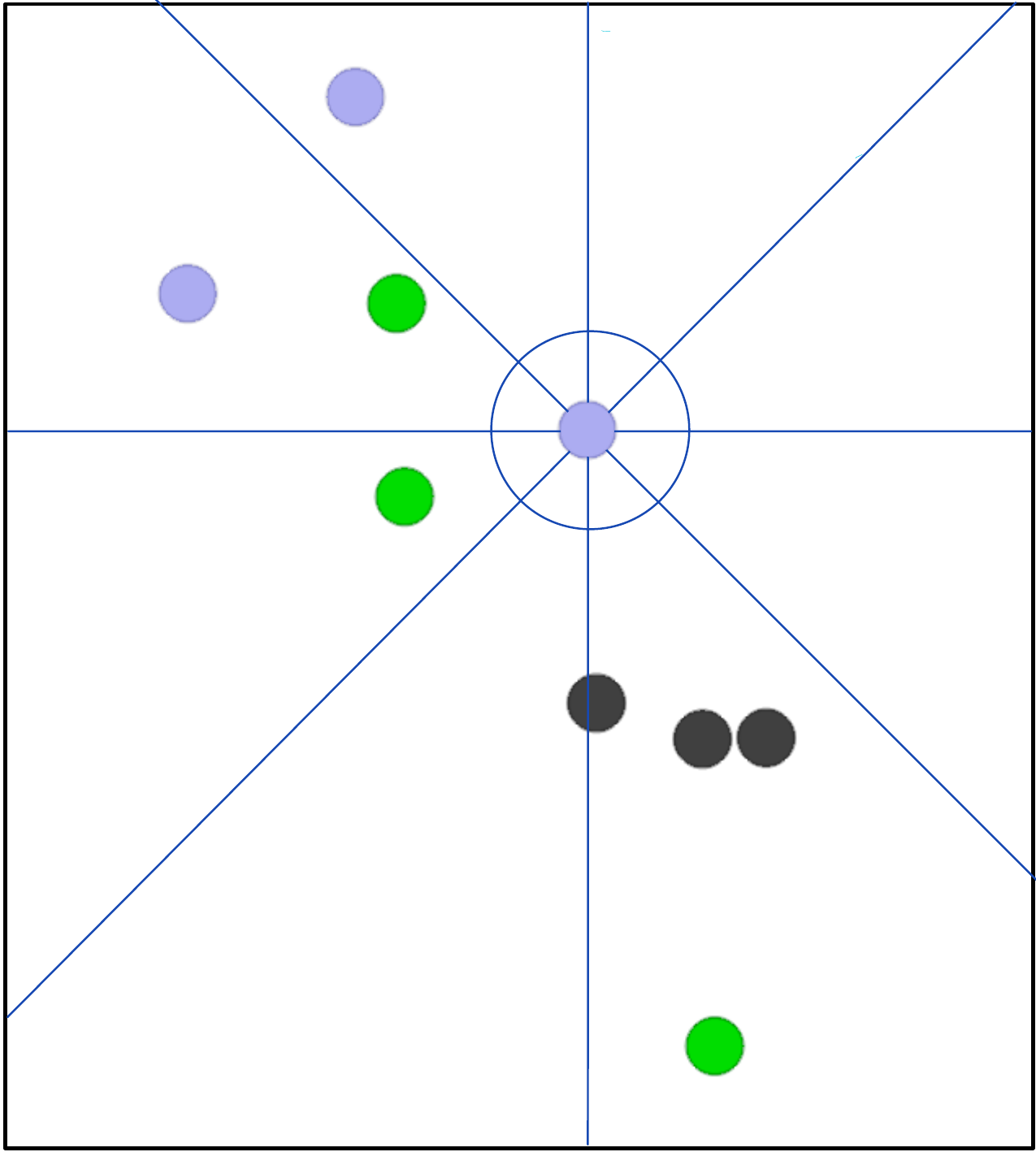}
    \caption{Example of the spatial clusters created by the 9 relations chosen for the continuous target task. Relations are shown with respect to one entity.}
    \label{fig:octagonal}
\end{figure}
For the continuous domain, we ran our experiment with default relations as described above, apart from the aligned relation, as the determination of that relation is not concise in the continuous domain. Similarly to our ablation studies on the discrete domain, we tested the addition of more detailed relations on the continuous domain. In this case, we added relations describing octagonal directions, plus adjacency, as depicted in Figure~\ref{fig:octagonal}. We found that the addition decomposes the state in more fine-grained areas and improves asymptotic performance as shown in Figure~\ref{fig:continuous_results}. This again highlights the trade-off between having a compact, more efficient representation and optimal performance.

\subsection*{Training Procedure and Hyperparameter}
The experiments were conducted on a workstation equipped with four NVIDIA RTX A6000 GPUs (46 GB each) and an AMD EPYC 7452 32-core Processor CPU (64 cores) with 251 GB RAM. The software environment consisted of Ubuntu 20.04.6, Python 3.9 and CUDA 12.4. The Python packages needed can be found in the \texttt{requirements.txt} file in our code base at https://github.com/sharlinu/MARC. 

Our implementation of MARC follows Algorithms \ref{alg:training} and \ref{alg:update}.
To find the optimal hyperparameters, we engaged in a random search for MARC on the complex task of LBF-10x10-4a-4f-coop, selecting the hyperparameters that perform best in terms of asymptotic performance. These hyperparameters, documented in Table~\ref{tbl:hyperparams}, are then applied consistently across all scenarios and are summarized in the following training procedure: We add each environment transition to a buffer of 1e5 transitions. The performance of MARC is robust to changes in the buffer length, which is why we chose a reduced buffer length to save memory. After every 100 environment steps, we perform 4 updates for both critics and policies using a batch size of 1024 samples. These updates are conducted using the Adam optimizer \cite{Adam} with a learning rate of 0.001 for both networks. The hidden dimension for all networks is uniformly set to 128, and the discount factor $\gamma$ is set at 0.99. Subsequently, we update the target networks using soft actor-critic updates, employing an update rate $\tau$ of 0.001. Additionally, the temperature parameter $\alpha$ is set to 0.05. For the policy network, we employ a simple 3-layer dense network. For the critic, we use a single-layered encoder to generate a dense feature representation out of the sparse entity features. These entity embeddings are then passed through a single-layered, shared R-GCN module. To output Q-values, we pass these entity embeddings through a max-pooling layer, generating a single vector representation that can be fed into a dense layer. Post concatenation with all agents' actions, these outputs traverse a final 2-layer dense network, unique to each agent.

\begin{algorithm}
\caption{Pseudocode for Multi-Agent Relational Actor-Critic}
\label{alg:training}
\begin{algorithmic}[1]
\STATE Initialize the environment and replay buffer $D$
\STATE $T_{\text{update}} \gets 0$
\FOR{$t = 1$ to $\text{num episodes}$}
    \STATE Reset environment and get initial observation $o_i$ for each agent $i =1,\dots, N$
    \WHILE{$\text{num}_{\text{steps}} < \text{max}_{\text{steps}}$ or episode $\neq$ terminated}
        \FOR{each agent $i$}
            \STATE Select action $a_{i} \sim \pi_{\theta_i}(\cdot|o_{i})$
            \STATE Do action and receive next observation $o_{i}'$ and reward $r_{i}$
            \STATE Store transitions $(o_i, a_i, r_i, o_i')$ in $D$
            \STATE $o_i \gets o_i'$
        \ENDFOR
        \STATE $T_{\text{update}} \gets T_{\text{update}} + 1$
        \IF{$T_{\text{update}} \geq \text{min}_{\text{update\_steps}}$}
            \STATE Sample a subset $B$ random transitions from $D$
            \FOR{$j = 1$ to $\text{num network updates}$}
                \STATE \text{UpdateCritics}($B$)
                \STATE \text{UpdatePolicies}($B$)
            \ENDFOR
            \STATE Soft update target parameters for all agents $i$:
            \STATE $\bar{{\psi_i}} \gets \tau \bar{\psi}_i + (1 - \tau){\psi_i}$
            \STATE $\bar{\theta_i} \gets \tau \bar{\theta}_i + (1 - \tau)\theta_i$
            \STATE $T_{\text{update}} \gets 0$
        \ENDIF
    \ENDWHILE
\ENDFOR
\end{algorithmic}
\end{algorithm}

\begin{algorithm}
\caption{Update Functions for Critic and Policies}
\label{alg:update}
\begin{algorithmic}[1]
\STATE \textbf{Function} UpdateCritics ($B$)
    \FOR{all agents $i= 1,\ldots,N$ and all transitions $b \in B$, in parallel}
        \STATE Calculate $Q_{\psi_i}(o^b_i, a^b_i)$
        \STATE Calculate ${a'_i}^b \sim \pi_{\bar{\theta}_i}({o'_i}^b)$ using target policy
        \STATE Calculate $Q_{\bar{{\psi_i}}}({o'_i}^b, {a'}^b)$ using target critic
    \ENDFOR
    \STATE Update critics by minimizing the joint regression loss $\mathcal{L}_Q(\psi)$
\STATE
\STATE \textbf{Function} UpdatePolicies ($B$)
    \FOR{all agents $i= 1,\ldots,N$ and all transitions $b \in B$, in parallel}
        \STATE Keep $o^b_i$ and discard the rest of the transition
        \STATE Sample new actions $a^b_i \sim \pi_{\theta_i}(o_i^b)$ for each agent
        \STATE Calculate $Q_{\psi_i}(o_i^b, a^b)$ using newly sampled actions
        \STATE Update policies using $\nabla_{\theta_i} J(\pi_{\theta_i})$
    \ENDFOR
\end{algorithmic}
\end{algorithm}

\begin{table*}
\vskip 0.15in
\begin{center}
\begin{tabular}{l||c|c}
\toprule
Parameter & MAAC/GA-AC & MARC \\
\midrule
\text { Dense layers in policy network} & 2 $\mid$  \textbf{3} & 2 $\mid$  \textbf{3} \\
\text { Dense layers in critic head } & \textbf{2} & \textbf{2} \\
\text { R-GCN layers} & - & 1 $\mid$  \textbf{2} $\mid$  3 \\
\text { Entity embedding hidden dimension} & - & 8 $\mid$  16 $\mid$  32 $\mid$  \textbf{48} $\mid$  64 $\mid$  128  \\
\text { Discount }($\gamma$) & \textbf{0.99} & \textbf{0.99} \\
\text { Replay buffer length } & $10^5$ $\mid$  $\mathbf{10^6}$ & $\mathbf{10^5}$ $\mid$  $10^6$ \\
\text { Critic learning rate}  & \textbf{0.001} $\mid$ 0.005 & \textbf{0.001} $\mid$ 0.005 \\
\text { Policy learning rate} & \textbf{0.001} $\mid$ 0.005 & \textbf{0.001} $\mid$ 0.005  \\
\text { Critic hidden dimension } & 64 $\mid$  \textbf{128} $\mid$  256 & 64 $\mid$  \textbf{128} $\mid$ 256 \\
\text { Policy hidden dimension } & 64 $\mid$  \textbf{128} $\mid$  256 & 64 $\mid$  \textbf{128} $\mid$ 256 \\
\text { Attention heads} & 1 $\mid$  2 $\mid$  \textbf{4}  $\mid$ 6 & -\\
\text { Batch size} & 512 $\mid$  \textbf{1024} & 512 $\mid$  \textbf{1024} \\
\text { Entropy coefficient }($\alpha$) & \textbf{0.01} & \textbf{0.01} \\
\text { Nonlinearity } & \textbf{LeakyReLU } & \textbf{LeakyReLU } \\
\text { Soft actor update rate }($\tau$) & 0.001 $\mid$  \textbf{0.005} $\mid$  0.01 & \textbf{0.001} $\mid$  0.005 $\mid$  0.01 \\
\text { Update interval in steps } & \textbf{100}   & \textbf{100}  \\
\text { Number of updates } & \textbf{4} & \textbf{4} \\
\text { Reward normalization } & False $\mid$ \textbf{True} &  False $\mid$ \textbf{True} \\
\bottomrule
\end{tabular}
\end{center}
\vskip -0.1in
\caption{Hyperparameter selection for MAAC, GA-AC and MARC. The table shows the range of evaluated hyperparameters, with the selected ones in bold.}
\label{tbl:hyperparams}
\end{table*}

% \begin{center}
%     \begin{table}[h] 
% \vskip 0.15in
% \begin{center}
% \begin{small}
% \begin{tabular}{l||c|c|c}
% \toprule
% Parameter & MAA2C & MAPPO & QMIX  \\
% \midrule
% \text { Hidden dimension } & 128 & 128 & 64 \\
% \text { Learning rate}  & 0.0005  & 0.0003 & 0.0003 \\
% \text { Parameter Sharing} & True  &  True  &  True \\
% \text { Policy Network} & Recurrent  &  FC  &  Recurrent \\
% \text { Reward standardization } & True &  False & True \\
% \text { Entropy coefficient } & 0.01 & 0.001 & - \\
% \text { Evaluation epsilon } & - & - & 0.5 \\
% \text { Epsilon anneal } & - & - & 200.000 \\
% \text { Soft target update } & 0.01 & 0.01  & 0.01 \\
% \text { n-step } & 10 & 5 & 10 \\ 
% \bottomrule
% \end{tabular}
% \end{small}
% \end{center}
% \vskip -0.1in
% \caption{Hyperparameter selection for MAA2C, MAPPO and QMIX (discrete). The table shows the selected parameters as proposed by \citet{papoudakis2020benchmarking}.}
% \label{tbl:hyperparams}
% \end{table}
% \end{center}

\subsection*{Baseline Implementations}

In the following, we list the code bases used to reproduce and test the baseline algorithms:
\begin{itemize}
    \item MAA2C, MAPP0 (discrete), QMIX (discrete): \url{https://github.com/uoe-agents/epymarl}
    \item InforMARL, MAPPO (continuous), QMIX (continuous): \url{https://github.com/nsidn98/InforMARL}
    \item GA-AC and MAAC : \url{https://github.com/shariqiqbal2810/MAAC}
\end{itemize}

Epymarl \cite{papoudakis2020benchmarking} is a thoroughly tuned and benchmarked codebase for MAPPO, MAA2C and QMIX that we used as baseline implementation for discrete tasks. We also leverage the comprehensive hyperparameter search conducted by \citet{papoudakis2020benchmarking}, adopting the optimal hyperparameters and architectures identified by the authors for the LBF environment and applying them consistently across our discrete tasks. Equally, InforMARL \cite{informal} was designed for and thoroughly tested on the continuous target task, using QMIX and MAPPO as their baselines. We therefore took their implementation and optimized hyperparameters for the continuous target task. Hence, for MAA2C, QMIX, MAPPO, and InforMARL we refer to paper and code base, where selected hyperparameters along with the matching training procedure are clearly and thoroughly documented. MAAC and GA-AC were not tested on the tasks that we used in our experiments so we conducted a random hyperparameter search for MAAC/GA-AC on the LBF-10x10-4a-4f-coop task, evaluated on asymptotic performance, and report the range and selection along with MARC in Table~\ref{tbl:hyperparams}. We note that GA-AC is based on the MAAC architecture with an additional hard-attention layer, so the implementation details for MAAC and GA-AC are equivalent. 

\subsection*{Alternative MARL Backbone}
\begin{figure}[h]
    \centering
    \includegraphics[width=\linewidth]{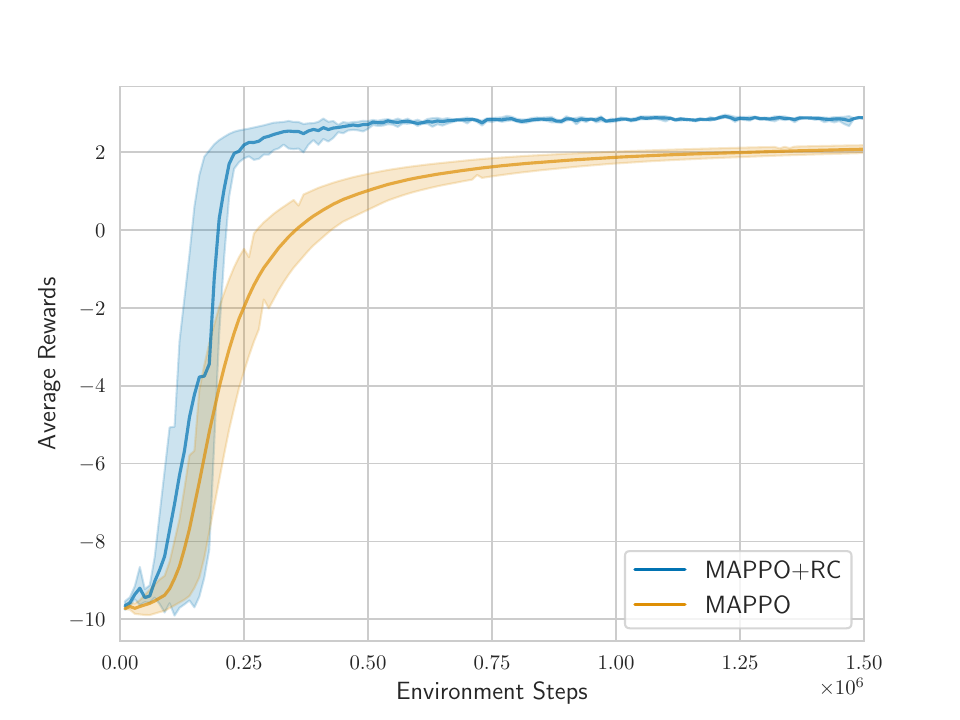}
    \caption{Mean average performance and $95\%$ confidence interval for a test version of CPP on a 7x7 grid with 1 picker agent, 1 delivery agent, 1 box and 1 goal locations. MAPPO denotes the original algorithm and MAPPO+RC the combination of MAPPO with a relational critic. For each model, we run 3 random seeds.} 
    \label{fig:mappo}
\end{figure}

In principle, relational abstraction is implemented as an observation encoder independent of the MARL algorithm itself. We therefore test this hypothesis by combining the relational observation encoder detailed in this paper with MAPPO \cite{MAPPO}, a strong and popular baseline algorithm. We compare our implementation with the original MAPPO implementation, setting the hyperparameters to be the same for a fair comparison. As seen in Figure~\ref{fig:mappo}, combining the observation encoder in the MAPPO critic architecture, we find an improvement in sample efficiency and asymptotic performance for the considered task.  

\subsection*{Invariances of the State Abstraction}

\begin{figure*}
\begin{subfigure}[t]{.24\linewidth}
    \centering
    \includegraphics[width =\linewidth]{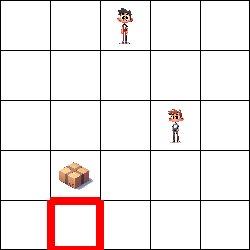}
\end{subfigure}
\begin{subfigure}[t]{.24\linewidth}
    \centering
    \includegraphics[width=\linewidth]{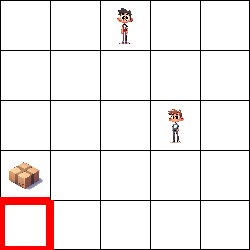}
\end{subfigure}
\begin{subfigure}[t]{.24\linewidth}
    \centering
    \includegraphics[width =\linewidth]{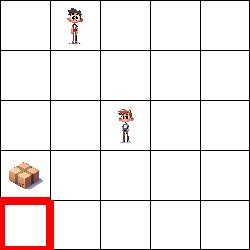}
\end{subfigure}
\begin{subfigure}[t]{.24\linewidth}
    \centering
    \includegraphics[width =\linewidth]{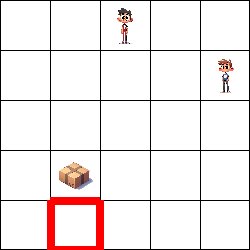}
\end{subfigure}
  \centering
  \caption{Examples of states that are considered to be equivalent in our critic architecture: translation shifts do not affect the representation if the relative spatial structure remains the same.}   \label{fig:abstraction_example}
\end{figure*}

By using graph neural network computations, we inherently benefit from its invariance to the order of the input elements \citep{hamilton2017inductive}. This is a very desirable property, as there is no natural ordering between the objects in an environment. Our constructed state mapping aggregates states together based on their relative spatial similarity. By fully removing the absolute information on positions and distance between entities, we not only reduce the state complexity but also induce a translation-invariant representation of the state. Translation shifts of the absolute positions of the environment objects do not influence the relative positioning of the environment objects, leaving the final structured representation unchanged. To demonstrate this, we give an example of what observations would be considered equal in the critic network in Figure~\ref{fig:abstraction_example}. 

% This becomes obvious when observing that the only component in the entity feature updates that take in absolute positions and could therefore be affected by shifts, is the computation of the relations $r \in \gR$. 
% Hence, we show the translation invariance on the example of one of the used relations: the left relation $\text{left}(u,v) \leftarrow x_u < x_v$.

Formally, we consider the translation operator $T_{(a,b)}$ that shifts the position of elements of a function $f$ by $(a,b)$, i.e. $(T_{a,b} f)(x,y) = f(x-a, y-b)$. Translation invariance means that
$$T_{(a,b)} f(x,y) = f(x,y),$$
i.e. the translation of the input leaves its output unchanged.
In our case, the only dependence on the position of an environment object $(x,y)$ lies in the construction of our edge set via our relational rules. More specifically, the relational rules are binary functions that depend on the absolute position of the entities $u$ and $v$ they are comparing. To demonstrate this in an example, we can rewrite the left relation as:
\begin{align*}
    r(u(x,y),v(x,y)) = \begin{cases} 
1 & \text{if } x_u < x_v \\ 
0 & \text{otherwise}.
\end{cases} 
\end{align*}

Applying the translation operator shifts the position of entities $u$ and $v$. So the translated relational rule can be written as:

\begin{align*}
r(T(u), T(v)) &= \begin{cases} 
1 & \text{if } x_u -a  < x_v - a \\ 
0 & \text{otherwise} 
\end{cases} \\
     &= \begin{cases} 
1 & \text{if } x_u < x_v \\ 
0 & \text{otherwise}. 
\end{cases}
\end{align*}
This ultimately implies translation invariance $r(u,v) = r(T(u), T(v))$, and can be shown for all spatial relational rules we employ.

\subsection*{Different Graph Architectures} \label{sec:graphs}
% All employed graph architectures can be written in the general form
% \begin{align*}
%  z_{i}'= update \left(z_{j}, \bigoplus_{j \in \gN_{i}}a(z_i,z_j) \psi \left(z_{j}\right)\right),
% \end{align*}
% where $\phi$ and $\psi$ are learnable and differentiable function, $a(z_i, z_j)$ is a coefficient determining the importance of entity$j$ to the representation of entity$i$ and $\bigoplus$ is a non-parametric, permutation-invariant function. 
% Such a graph is formally defined as $\gG = (\gV, \gE, \gR, Z)$, where $\gV$ is the set of entities, $\gE$ is the set of edges, $\gR$ is the set of relations and $Z$ contains entity features.
To introduce a GAT-layer \citep{GAT2018}, we construct a complete binary graph $\gG' = (\gV, \gE', Z')$, where, as before, $\gV$ consists of N agents and M objects. However, for the GAT implementation, we drop all relation types and fully connect all entities with each other. To add back the spatial information, we enrich the feature matrix $Z' \in \mathbb{R}^{(d+2) \times |\gV|}$ with two additional dimensions for the coordinates of the entities. The update of the entity features is as follows:
\begin{align*}
a(z_i, z_j) &= \operatorname{softmax} \left(\operatorname{LeakyReLU} \left( q^T W z_i + k^T W z_j \right) \right), \\
 z_{i}' &=\sigma \left(\sum_{j \in \gN(i)\cup \{i\} } a(z_i, z_j) W z_{j} \right)
\end{align*}
%{\sum_{k \in \mathcal{N}_i}\left(\operatorname{LeakyReLU}\left(\overrightarrow{\mathbf{a}}^T\left[\mathbf{W} \vec{h}_i \| \mathbf{W} \vec{h}_k\right]\right)\right)}
where $q,k \in \mathbb{R}^{d'}$ and $W \in \mathbb{R}^{d' \times d}$ are weight vectors and matrix, respectively. $ a(z_i, z_j)$ is a learnable, self-attention weight that implicitly computes the importance of entity $j$ to the representation of entity $i$. That way, we can learn the importance of connection between the agents and other objects in the environment. 

One can combine the GAT architecture with an R-GCN by following \citet{RGAT2019} in applying attention weights to edges of a heterogeneous graph $ \gG = (\gV, \gE,\gR, Z)$, as it is defined earlier. This yields the following update:
\begin{align*}
a_r &(z_i, z_j) \\ &= \operatorname{softmax} \left(\operatorname{LeakyReLU} \left( q^T_r W_r z_i + k^T_r W_{r} z_j \right) \right), \\
 z_{i}'&=\sigma\left(\sum_{r \in \gR} \sum_{j \in \gN_{r}(i)} a_r(z_i, z_j) W_{r} z_{j}\right).
\end{align*}
where $q_{r},k_{r} \in \mathbb{R}^{d'}$ and $W_r \in \mathbb{R}^{d' \times d}$ are relation-specific weight vectors and matrix, respectively. $a_r(z_i, z_j)$ are, as before, learnable self-attention weights but now depend on the specific relation type. The $\operatorname{softmax}(\cdot)$ is now computed across all connecting entities irrespective of their relations. That means that $a_r(z_i, z_j)$ implicitly computes the importance of entity $j$ to the representation of entity $i$ under relation type $r$, compared to all incoming connections to entity $i$. The difference between the R-GAT and R-GCN update is that in R-GCN, each neighboring entity has equal importance and is simply weighted with a relation-specific normalizing constant, i.e. $a_r(z_i, z_j)  = |\gN_{r}(i)|^{-1}$.

% \section{Reproducibility Checklist Comments}
% Following we want to elaborate on the comments we answered with partial in the reproducibility checklist
% \begin{itemize}
%     \item \textit{This paper specifies the computing infrastructure used for running experiments (hardware and software), including GPU/CPU models; amount of memory; operating system; names and versions of relevant software libraries and frameworks} - we train
%     \item \textit{This paper lists all final (hyper-)parameters used for each model/algorithm in the paper’s experiments} - For the hyper-parameter search we conducted, we clearly list all selected hyper-parameters for MARC, MAAC and GA-AC. 
% \end{itemize}

% \bibliography{aaai25}

\end{document}